\def\paperID{6305}
\def\confName{ICCV}
\def\confYear{2025}

\def\paperTitle{\om{}: Towards Referring Expression Segmentation in the Wild\\via Synthetic Data}

\def\authorBlock{
    Dong-Hee Kim \qquad
    Hyunjee Song \qquad 
    Donghyun Kim \\ 
    Korea University \\
    {\tt\small \{dongheekim, kelly062001, d\_kim\}@korea.ac.kr}
}

\newif\ifreview 
\newif\ifarxiv \newcommand{\arxiv}{\arxivtrue}
\newif\ifcamera 
\newif\ifrebuttal 

\arxiv

\pdfoutput=1
\documentclass[10pt,twocolumn,letterpaper]{article}
\ifreview \usepackage[review]{cvpr} \fi
\ifarxiv \usepackage[pagenumbers]{cvpr} \fi
\ifrebuttal \usepackage[rebuttal]{cvpr} \fi
\ifcamera \usepackage{cvpr} \fi

\usepackage{graphicx}	
\usepackage{amsmath}	
\usepackage{amssymb}	
\usepackage{booktabs}
\usepackage{times}
\usepackage{microtype}
\usepackage{epsfig}
\usepackage{caption}
\usepackage{float}
\usepackage{placeins}
\usepackage{color, colortbl}
\usepackage{stfloats}
\usepackage{enumitem}
\usepackage{tabularx}
\usepackage{xstring}
\usepackage{multirow}
\usepackage{xspace}
\usepackage{url}
\usepackage{subcaption}
\usepackage{xcolor}
\usepackage[hang,flushmargin]{footmisc}
\usepackage{pifont}
\usepackage{makecell}
\usepackage{pgfplots}
\usepackage{tcolorbox}
\usepackage{titletoc}
\usepackage{wrapfig}

\definecolor{cvprblue}{rgb}{0.21,0.49,0.74}

\ifcamera \usepackage[accsupp]{axessibility} \fi

% ...

\newcommand{\nbf}[1]{{\noindent \textbf{#1.}}}

\newcommand{\cmark}{\ding{51}\xspace}%
\newcommand{\xmark}{\ding{55}\xspace}

\newcommand{\tabincell}[2]{\begin{tabular}{@{}#1@{}}#2\end{tabular}}
\newcommand{\om}{SynRES}

\newcommand{\tablestyle}[2]{\setlength{\tabcolsep}{#1}\renewcommand{\arraystretch}{#2}\centering\footnotesize}
\newlength\savewidth\newcommand\shline{\noalign{\global\savewidth\arrayrulewidth
		\global\arrayrulewidth 1pt}\hline\noalign{\global\arrayrulewidth\savewidth}}

\newcommand{\dt}[1]{\fontsize{5pt}{0.1em}\selectfont (#1)}

\newcommand{\supp}{supplemental material\xspace}
\ifarxiv \renewcommand{\supp}{appendix\xspace} \fi

\newcommand{\wildres}{WildRES\xspace}
\newcommand{\ours}{SynRES\xspace}

\newcommand{\R}[1]{{%
    \textbf{%
        \ifstrequal{#1}{1}{\textcolor{magenta}{R#1}}{%
        \ifstrequal{#1}{2}{\textcolor{teal}{R#1}}{%
        \ifstrequal{#1}{3}{\textcolor{orange}{R#1}}{%
        \ifstrequal{#1}{4}{\textcolor{teal}{R#1}}{%
                           \textcolor{cyan}{R#1}%
        }}}}%
    }%
}}

\usepackage{xr-hyper}

\makeatletter
\newcommand*{\addFileDependency}[1]{
  \typeout{(#1)}
  \@addtofilelist{#1}
  \IfFileExists{#1}{}{\typeout{No file #1.}}
}

\makeatother
\newcommand*{\myexternaldocument}[1]{
    \externaldocument{#1}
    \addFileDependency{#1.tex}
    \addFileDependency{#1.aux}
}

\definecolor{cvprblue}{rgb}{0.21,0.49,0.74}
\usepackage[pagebackref,breaklinks,colorlinks,allcolors=cvprblue]{hyperref}
\usepackage[capitalize]{cleveref}
\usepackage{xcolor}
\crefname{section}{Sec.}{Secs.}
\crefname{table}{Table}{Tables}
\crefname{figure}{Fig.}{Figs.}

\ifarxiv \crefname{appendix}{App.}{Apps.}
\else \crefname{appendix}{Suppl.}{Suppls.} \fi

\myexternaldocument{_supplementary}

\begin{document}
\title{\paperTitle}
\author{\authorBlock}


\twocolumn[{%
\renewcommand\twocolumn[1][]{#1}%
\maketitle 
\vspace{-5mm}
\begin{center} 
\centering 
\includegraphics[width=1\textwidth]{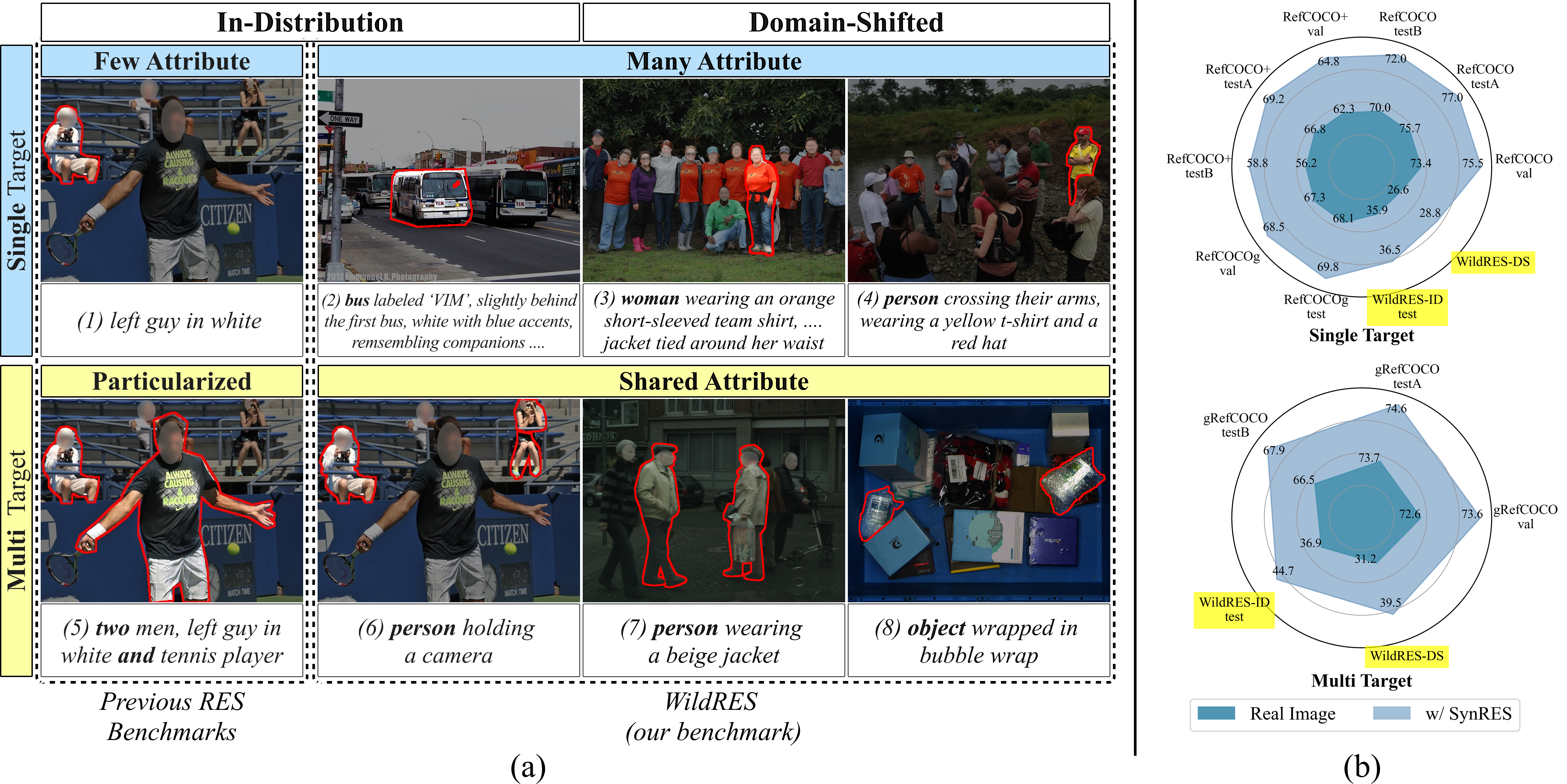} 
\vspace{-6mm}
\captionof{figure}{(a) Comparison of existing referring expression segmentation (RES) benchmarks with our proposed benchmark, \wildres, which demands advanced complex reasoning and scene understanding. Previous RES datasets~\cite{yu2016refcocorefcoco+, liu2023gres} primarily emphasize single-target focus with short queries (1) or distinctive multi-target queries (5) within a similar domain (\eg, images from COCO~\cite{caesar2018coco}). These constraints limit the evaluation of complex queries and generalization ability of RES models. We propose a new benchmark \wildres, which improves single-target expressions with diverse attributes (2)-(4) and refines multi-target ones with shared traits (non-distinctive) and concise phrasing (6)-(8). In addition, \wildres includes in-distribution (col. 2) and domain-shifted subsets (cols. 3, 4) to evaluate generalizability. (b) The state-of-the-art RES method (LISA~\cite{lai2024lisa}) experiences notable performance degradation on our benchmark (highlighted in yellow), which requires advanced reasoning and generalization capability. Our novel synthetic data generation (\om{}) enhances the model's compositional reasoning ability.}
\label{fig:wildseg_comparison}
\end{center}%
}]


%
    
\begin{abstract}
    Despite the advances in Referring Expression Segmentation (RES) benchmarks, their evaluation protocols remain constrained, primarily focusing on either single targets with short queries (containing minimal attributes) or multiple targets from distinctly different queries on a single domain. This limitation significantly hinders the assessment of more complex reasoning capabilities in RES models.
    We introduce \wildres, a novel benchmark that incorporates long queries with diverse attributes and non-distinctive queries for multiple targets. This benchmark spans diverse application domains, including autonomous driving environments and robotic manipulation scenarios, thus enabling more rigorous evaluation of complex reasoning capabilities in real-world settings. 
    Our analysis reveals that current RES models demonstrate substantial performance deterioration when evaluated on \wildres. To address this challenge, we introduce \ours, an automated pipeline generating densely paired compositional synthetic training data through three innovations: (1) a dense caption-driven synthesis for attribute-rich image-mask-expression triplets, (2) reliable semantic alignment mechanisms rectifying caption-pseudo mask inconsistencies via Image-Text Aligned Grouping, and (3) domain-aware augmentations incorporating mosaic composition and superclass replacement to emphasize generalization ability and distinguishing attributes over object categories.
    Experimental results demonstrate that models trained with \ours achieve state-of-the-art performance, improving gIoU by 2.0\% on \wildres-ID and 3.8\% on \wildres-DS.
    Code and datasets are available at \url{https://github.com/UTLLab/SynRES}.
\end{abstract}

\vspace{-3mm}
\section{Introduction}
\label{sec:intro}

Recent advancements in Large Multimodal Models (LMMs)~\cite{liu2024vqa1, liu2024vqa2} and foundational segmentation models~\cite{kirillov2023segment, ravi2025sam2} have significantly enhanced language-based image segmentation by improving both open vocabulary capability and sentence understanding for Referring Expression Segmentation (RES)~\cite{lai2024lisa, xia2024gsva, ren2024pixellm, rasheed2024glamm, zhang2024psalm, chen2024sam4mllm, zhang2024evf}. These models surpass the constraints of traditional RES approaches~\cite{wang2022cris, liu2023gres, yang2022lavt, zhao2023vpd, xu2023etris, zou2023seem}, which rely on completed image-mask-expression triplet datasets~\cite{yu2016refcocorefcoco+, mao2016refcocog, liu2023gres}, by leveraging diverse data sources such as semantic segmentation~\cite{caesar2018coco, zhou2017ade20k, neuhold2017mapillary, chen2014pascalpart, he2022partimagenet}, referring image segmentation~\cite{yu2016refcocorefcoco+, mao2016refcocog, liu2023gres}, and visual question answering (VQA)~\cite{liu2024vqa1, antol2015vqa, liu2024vqa2}. This broader data integration allows these models to effectively interpret diverse expressions and operate robustly in real-world scenarios.

Furthermore, numerous RES benchmarks~\cite{rohrbach2016refclef,yu2016refcocorefcoco+,mao2016refcocog,liu2023gres} have emerged to evaluate models' compositional reasoning and generalization capabilities. Notably, gRefCOCO~\cite{liu2023gres} introduces multi-target or no-target expressions to further assess models' advanced reasoning abilities. Nevertheless, existing benchmarks predominantly concentrate on expressions with minimal attributes~\cite{yu2016refcocorefcoco+, rohrbach2016refclef} or distinctive (countable) expressions~\cite{liu2023gres} within the homogeneous domain of COCO~\cite{kazemzadeh2014refcoco}, as illustrated in Fig.~\ref{fig:wildseg_comparison}-(a. 1, 5). This narrow focus substantially constrains the assessment of complex reasoning capabilities required in real-world scenarios. 
In this paper, we introduce \textbf{\wildres}, a new test benchmark designed to reflect advanced reasoning and real-world complexity. \wildres incorporates three new elements as shown in Fig.~\ref{fig:wildseg_comparison}-(a. 2-4, 6-8): (1) explicit longer queries containing many attributes, (2) shared (non-distinctive) attribute expressions for multiple targets, and (3) domain-shifted images from challenging environments including densely populated scenes~\cite{shao2018crowdhuman}, autonomous driving contexts~\cite{Cordts2016Cityscapes}, and robotic applications~\cite{mitash2023armbench}. Fig.~\ref{fig:wildseg_comparison}-(b) shows the existing RES model (\eg,~\cite{lai2024lisa}) exhibits substantial performance deterioration when evaluated on \wildres.




A natural approach would be to acquire large-scale training data encompassing similar expressions and domain-shifted imagery comparable to those in \wildres. However, obtaining such annotations would be prohibitively expensive. To overcome this limitation, we utilize a transfer learning approach leveraging automatically generated synthetic data with compositional queries. Previous works~\cite{wu2023diffumask, nguyen2023datasetdiffusion, yang2023freemask, ye2024seggen} generate synthetic data for semantic segmentation by utilizing a generative model conditioned on target masks, but it does not provide referring expressions necessary for RES. Pseudo-RIS~\cite{yu2024pseudo} generates synthetic text captions and pseudo-masks from SAM~\cite{ravi2025sam2} using the training dataset in COCO~\cite{kazemzadeh2014refcoco}. However, generated data is not specifically designed for multiple targets or non-distinctive queries, consequently resulting in an insufficiency of densely paired image-expression combinations. The fundamental questions of \emph{how} to generate effective and \emph{how} to optimally leverage synthetic data for complex queries with diverse attributes and domains remain underexplored. 

We propose \textbf{\ours}, which automatically generates densely paired synthetic referring expressions with many attributes and corresponding images and masks triplets via generative models. As illustrated in Fig.~\ref{fig:overall}, \ours produces images containing identical objects with heterogeneous attributes, which will subsequently be utilized to train models capable of discriminating between various attribute combinations.
\ours consists of three key steps: 1) generating distinctive synthetic expressions with different attributes for the same object and their corresponding images, 2) grouping semantically similar expressions and aligning them with corresponding synthetic masks, and 3) domain-aware image augmentation and semantic text augmentation to reduce domain gaps in each modality.  In step 1, we utilize real images and masks along with image captioning models like CoCa~\cite{yu2022coca} to create distinctive captions~\cite{yu2024pseudo}. Unlike minimal human expressions
these synthetic expressions incorporate rich attributes to improve performance under challenging scenarios (see Fig.~\ref{fig:augmentation}). These concatenated expressions are then used as input for Text-to-Image (T2I) models to generate synthetic images encompassing all expression features. Step 2 addresses potential inaccuracies in pseudo-masks of the generated images. 
To resolve such issues, we group related expressions and create segmentation masks aligned with these groups. Finally, in step 3, we enhance the synthetic image-expression-mask triplets through data augmentation techniques tailored for RES in the wild. These include mosaic augmentation for multi-target scenarios and superclass replacement text semantic augmentation to focus on detailed target descriptions rather than single-word references.

The proposed \om method significantly enhanced referring expression segmentation (RES) model performance across diverse benchmarks and architectural configurations. Specifically, LISA achieved substantial improvements of +2.0 and +2.8 $\text{gIoU}$ on \wildres-ID (same domain), and GSVA also demonstrated comparable gains through \om integration. The approach enhanced cross-domain generalization by 3.8--6.2 gIoU on \wildres-DS (domain shifted), consistently outperforming existing baselines on standard RES benchmarks. Comprehensive ablation studies underscored the critical importance of both textual and visual augmentation in \om.


\noindent
Our contributions are summarized as follows:
\begin{itemize}
    \item We propose \wildres, a novel benchmark for Referring Expression Segmentation in real-world scenarios, covering single-target many-attribute cases and  multi-target shared-attribute. \wildres includes diverse domains such as crowded scenes, autonomous driving, and robotics.
    \item We introduce \om{}, an automated pipeline for generating densely paired synthetic datasets with diverse attributes and precise pseudo-masks, enabling effective data augmentation without manual annotation.
    \item Experiments demonstrate that \ours is model-agnostic and enhances existing RES models, showing improvements in wild scenarios and classic benchmarks. 
\end{itemize}

\vspace{-1mm}
\section{Related Work}
\label{sec:related}

\subsection{Referring Expression Segmentation with Large Multimodal Models}
\label{subsec:related_RES_datasets}
The introduction of LISA~\cite{lai2024lisa}, which leveraged the visual-language capabilities of Large Multimodal Models (LMMs) for the Referring Expression Segmentation (RES) task using \texttt{<SEG>} tokens, has significantly influenced the development of advanced LMM-based RES models~\cite{xia2024gsva,ren2024pixellm,rasheed2024glamm,zhang2024psalm,chen2024sam4mllm,zhang2024evf}. Notably, GSVA~\cite{xia2024gsva} and PixelLM~\cite{ren2024pixellm} have improved multi-target segmentation accuracy by employing strategies such as multiple \texttt{<SEG>} tokens or introducing target refinement loss. To train these models effectively, large-scale datasets from diverse domains are utilized. Examples include semantic segmentation datasets~\cite{caesar2018coco,zhou2017ade20k,neuhold2017mapillary, chen2014pascalpart, he2022partimagenet}, classic RES datasets~\cite{yu2016refcocorefcoco+,mao2016refcocog}, and Visual Question Answering datasets~\cite{liu2024vqa1,antol2015vqa,liu2024vqa2}. These datasets are adapted for RES tasks using prompt-based transformations or integrated into training pipelines. Such adaptations enhance vocabulary comprehension and reasoning capabilities. Some models employ specialized datasets tailored for generalized RES tasks. For instance, GSVA~\cite{xia2024gsva} and SAM4LMM~\cite{chen2024sam4mllm} incorporate gRefCOCO~\cite{liu2023gres} into their training data. Other notable datasets include ReasonSeg~\cite{lai2024lisa} and GranD~\cite{rasheed2024glamm}, which are derived from automatically annotated SA-1B~\cite{kirillov2023segment} or GranD-f (based on Flickr-30K~\cite{young2014flickr}, RefCOCOg~\cite{mao2016refcocog}, and PSG~\cite{yang2022psg}). Existing models exhibit performance degradation when evaluated on our more challenging benchmark, \wildres. Our model-agnostic approach, \ours, significantly enhances the complex reasoning capabilities of these models.

\subsection{Synthetic Data Generation for Segmentation}
Initial efforts in synthetic dataset generation for image segmentation were driven by Generative Adversarial Networks (GANs)~\cite{zhang2021datasetgan,li2022bigdatasetgan}. With the emergence of generative diffusion models~\cite{ho2020denoising,song2021denoising}, recent research has increasingly focused on creating synthetic datasets for segmentation tasks. These methods generate images corresponding to specific semantic segmentation classes alongside paired pseudo-segmentation masks~\cite{nguyen2023datasetdiffusion,yang2023freemask}. Additionally, diffusion models have been used to synthesize masks and generate corresponding high-quality images for semantic segmentation~\cite{ye2024seggen}. These techniques have also been extended to instance and panoptic segmentation tasks~\cite{zhao2023xpaste,fan2024divergen,xie2024mosaicfusion,tu2025dreammask}. While existing methods focus on image-mask pairs, our approach generates densely aligned image-mask-text triplets (\eg, same objects containing diverse attributes, and its reliable pseudo-mask) specifically optimized for RES requirements.

\subsection{Data Augmentation for RES}
Multimodal data augmentation methods~\cite{hao2023mixgen,liu2023lembda,jin2024armada,wu2023biaug} have advanced feature learning by integrating text and image modalities. However, techniques specifically tailored for RES remain underexplored. To address this gap, NeMo~\cite{ha2024finding} employs a pretrained CLIP~\cite{radford2021clip} model to select similar yet non-redundant images for mosaic augmentation. Pseudo-RIS~\cite{yu2024pseudo} uses SAM~\cite{kirillov2023segment} to create pseudo masks and CoCa~\cite{yu2022coca} to generate unique image captions. This enables weakly supervised training that enhances performance even with annotated data. However, these techniques often fail to comprehensively address the interplay between image, text, and segmentation masks in the context of RES~\cite{ha2024finding}.  Our method systematically coordinates all three modalities through novel augmentation strategies that mitigate domain discrepancies while enhancing generalization capabilities and semantic consistency across transformations.

\vspace{-1mm}

\section{Referring Expression Segmentation in the Wild}

\begin{figure}[tp]
    \centering
    \includegraphics[width=\linewidth]{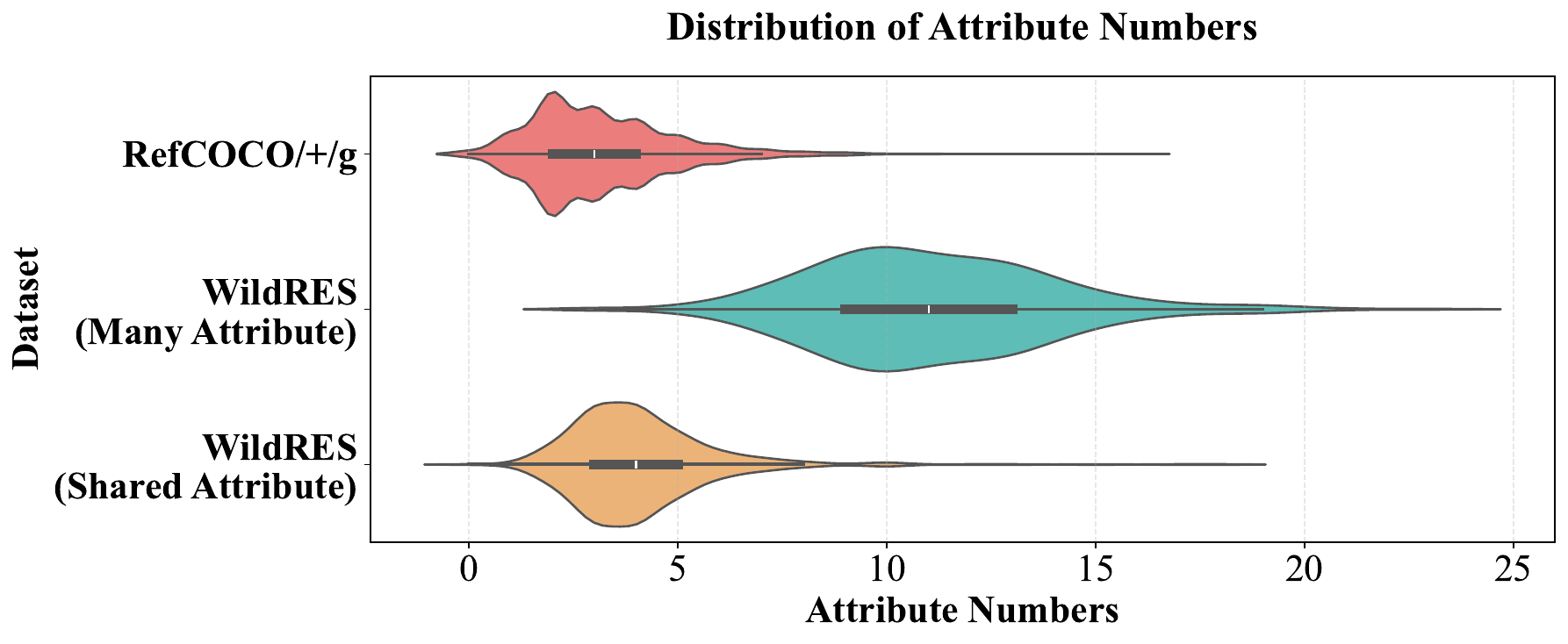}
    \vspace{-6mm}
    \caption{Number of attributions in classic RES datasets vs. \wildres. Using GPT-4o-mini~\cite{achiam2023gpt4o}, we counted the number of attributes in Table~\ref{tab:refer_attr}. Single-target expressions in \wildres often exceed 11 attributes to specify a target, while multi-target expressions have fewer attributes, similar to classic RES datasets.}
    \vspace{-4mm}
    \label{figs:wildseg_attr_num}
\end{figure}

\subsection{Problem Definition}

In RES, given an input image $\mathbf{x}_\text{img}$ and an input language query $\mathbf{x}_\text{txt}$, the goal is to produce the corresponding binary segmentation mask $\mathbf{M}$. In addition to this traditional RES, we aim to address the three key aspects: \romannumeral1) using referring expressions with diverse attributes to pinpoint a single precise target in Fig.~\ref{fig:wildseg_comparison}-(a, 2-4), \romannumeral2) identifying specific multiple targets with shared attributes in Fig.~\ref{fig:wildseg_comparison}-(a, 6-8), and \romannumeral3) extending these two aspects to other domains, as demonstrated in Fig.~\ref{fig:wildseg_comparison}-(a, 3, 4, 7, 8).

For multi-target scenarios, we select images containing at least three objects and design the referring expressions so that only a subset of these objects satisfies the specified attributes. Detailed descriptions emphasize unique characteristics of individual objects, compelling segmentation models to discern both shared and distinct features.

\vspace{-3pt}
\subsection {\wildres: A New RES Benchmark for Advanced Reasoning and Generalization}
\begin{figure*}[tp]
    \centering
    \includegraphics[trim=0mm 0mm 6mm 0mm, clip, width=\textwidth]{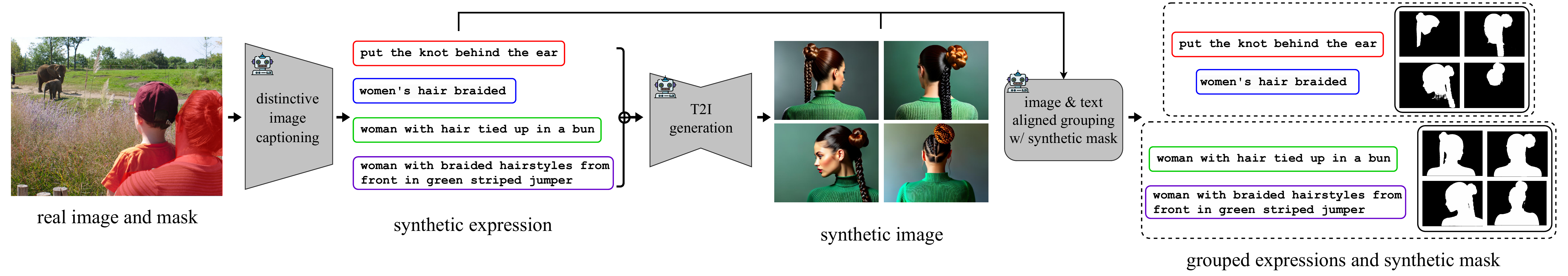}
    \vspace{-6mm}
    \caption{Overview of the step 1 and 2 in \om{}. The process begins by creating distinctive $n$ synthetic expressions for target objects from real images and masks. These expressions are then concatenated and input into a pre-trained text-to-image generative model to produce $m$ synthetic images. Finally, a grouping step is performed to generate reliable synthetic masks by associating appropriate $l$ segmentation masks with their corresponding expressions, yielding densely paired image-mask-expression triplets for objects with diverse attributes, thereby facilitating the learning of distinctive attribute combinations across objects.} 
    \label{fig:overall}
    \vspace{-4mm}
\end{figure*}

We introduce \wildres, a benchmark designed to evaluate segmentation models in complex, real-world contexts with 724 images and 941 expressions. \wildres does not contain the training set, only the validation and test set. \wildres includes both in-distribution images (\wildres-ID) sourced from MSCOCO~\cite{caesar2018coco} and domain-shifted images (\wildres-DS) representing unseen scenes from datasets such as CrowdHuman~\cite{shao2018crowdhuman}, Cityscapes~\cite{Cordts2016Cityscapes}, and ARMBench~\cite{mitash2023armbench}, all aimed at assessing model robustness in challenging environments.

\wildres is divided into two subsets: Many Attribute (MA) and Shared Attribute (SA). MA subsets comprise images that contain at least 5 objects, with each caption manually reviewed to ensure precise alignment with its corresponding segmentation mask. SA subsets focus on images featuring multiple, referable objects from the same class that exhibit distinct attributes, thereby evaluating fine-grained segmentation capabilities. In \wildres-ID, both subsets are present; in contrast, within \wildres-DS, datasets are categorized according to their dominant features—CrowdHuman, which contains numerous individuals, corresponds to the Many Attribute subset, while Cityscapes and ARMBench, featuring multiple objects with shared attributes, align with the Shared Attribute subset. The exact quantity for each image and expression constituting \wildres is described in detail in \supp~\ref{sec:appendix_wildres_info}.

Existing RES datasets often lack diversity in their attributes. To systematically analyze this limitation, we categorized referring expressions into 8 distinct attributions (more details of each attributions are in \supp~\ref{sec:attr_type}). Utilizing GPT-4o-mini~\cite{achiam2023gpt4o}, we classified referring expressions from these datasets into the predefined categories. Fig.~\ref{figs:wildseg_attr_num} illustrates the results of this analysis. We observed that expressions from traditional RES datasets are composed of simple and easily targetable samples, with a mean attribute count below 4. In contrast, \wildres's Many Attribute expressions exceed an average of 11 attributes, while maintaining a consistent number of attributes in expressions when referring to multiple targets sharing attributes. Detailed differences from previous datasets~\cite{liu2023gres, you2025pix2cap} are described in \supp~\ref{sec:wildres_diff}.


\vspace{-3pt}
\section{Our Method: \om{}}
\vspace{-3pt}
\label{sec:method}




We propose \om{}, a synthetic dataset generation and augmentation pipeline for challenging RES, as shown in Fig.~\ref{fig:overall}. 
As existing data generation approaches~\cite{yang2023freemask,yu2024pseudo} do not produce densely paired image-mask-expression triplets, they could be suboptimal for developing models capable of discriminating between objects based on diverse attribute queries extracted from synthetic data. \om{} generates densely paired distinctive synthetic expressions and their corresponding images in Step 1. We obtain reliable pseudo semantic segmentation masks with image-text aligned grouping in Step 2. Our synthetic dataset is subsequently enhanced with data augmentation techniques to train models with improved generalization capabilities in Step 3.

\vspace{-1mm}
\subsection{Step 1: Synthetic Distinctive Referring Expressions and Image Generation}
\vspace{-1mm}
\label{sec:step1}
We first aim to generate distinctive synthetic expressions with diverse attributes. Motivated by ~\cite{yu2024pseudo}, \om{} employs image captioning models to generate distinctive expressions $\{\mathbf{x}^\text{syn}_{\text{txt}, j}\}_{j=1}^{n}$ for individual referring objects from real image $\mathbf{x}^\text{real}_{\text{img}}$ and it's corresponding mask $\mathbf{M}$ as in Fig.~\ref{fig:overall}. 

For image synthesis, we construct composite prompts by aggregating the generated expressions to ensure comprehensive feature representation. For example, by combining the expressions ``\textit{put the knot behind the ear}'' and ``\textit{women's hair braided}'', the aggregated description ``\textit{put the knot behind the ear, women's hair braided}'' yields structured inputs with structured template selection per generation instance (see prompt templates in \supp~\ref{sec:t2i_prompt}). We leverage SANA~\cite{xie2024sana}, a text-to-image model that efficiently generates high-fidelity synthetic images $\{\mathbf{x}^\text{syn}_{\text{img}, i}\}_{i=1}^{m}$ with different $m$ seeds while maintaining visual-semantic alignment.

\vspace{-1mm}
\subsection{Step 2: Reliable Synthetic Mask Generation with Image-Text Aligned Grouping}
\vspace{-1mm}
\begin{figure}[tp]
    \centering
    \includegraphics[trim=6mm 9mm 4mm 8mm, clip, width=0.88\linewidth]{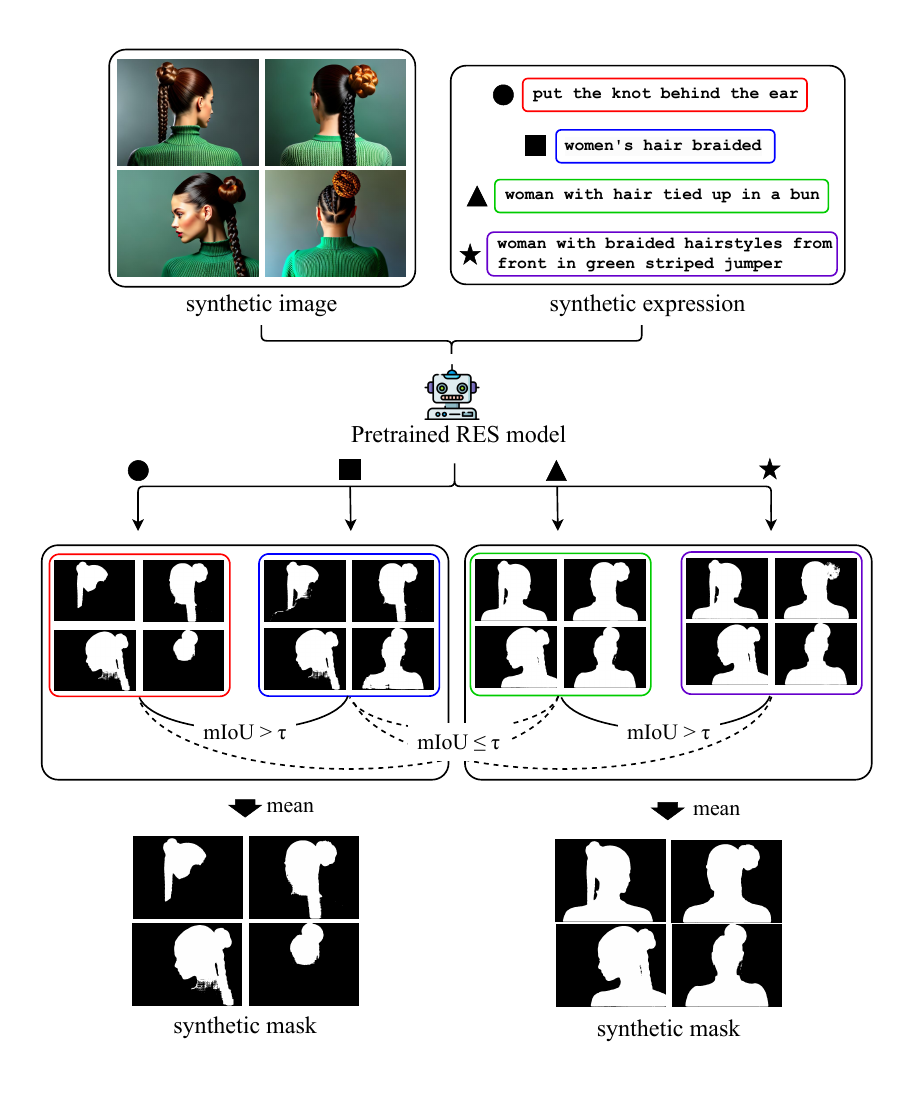}
    \vspace{-2mm}
    \caption{Image-text aligned grouping for reliable pseudo mask generation. Synthetic images and textual expressions are aligned through pseudo mask generation, binary conversion, and pairwise IoU-based clustering to form consensus groups. Final refined segmentation masks are computed via per-group averaging and thresholding, ensuring high-quality mask alignment.}
    \label{fig:grouping}
    \vspace{-5mm}
\end{figure}
\begin{figure*}[tp]
    \centering
    \includegraphics[trim=6mm 1mm 6mm 1mm, clip, width=\textwidth]{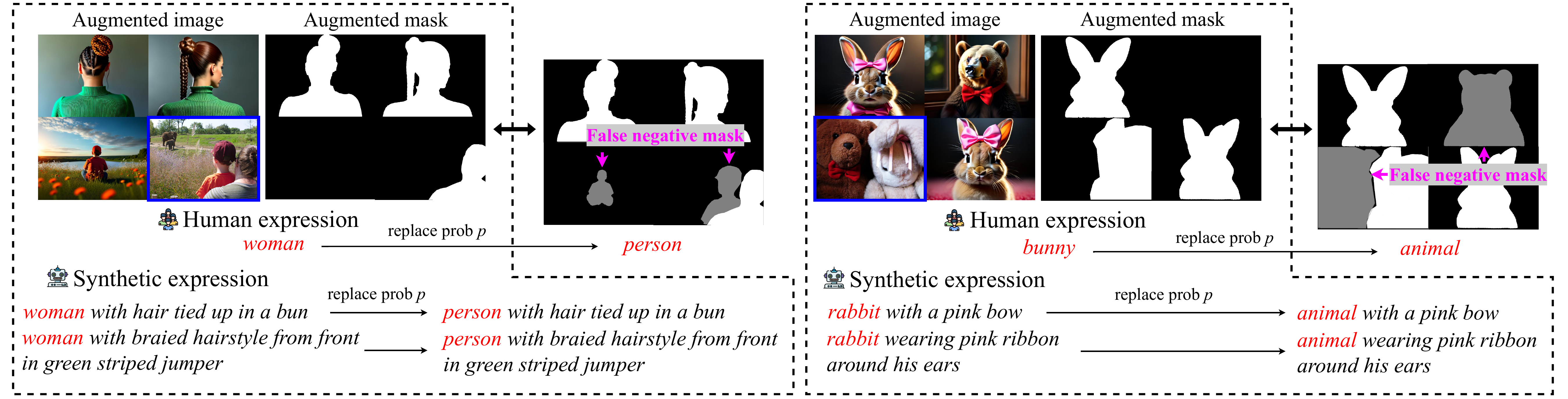}
    \vspace{-6mm}
    \caption{Augmented examples in the step 3. Mosaic augmentation is applied using synthetic images and masks containing one original real image (\textcolor{blue}{blue border}) and masks. Specific words (\eg, \textcolor{red}{woman}) are replaced with their superclass (\eg, \textcolor{red}{person}), a broader concept, with a probability of $p$, which could effectively mitigate model bias toward specific terminology and facilitate the learning of broader vocabulary associations.
    This substitution process may potentially produce false negative masks (\textcolor{Magenta}{magenta arrows}) when other objects belonging to the same superclass exist within the image. Although this frequently exists in human expressions (\eg, RefCOCO+), we manually mitigate this challenge by generating isolated single objects within our synthetic data. }
    \vspace{-2mm}
    \label{fig:augmentation}
\end{figure*}

After the step 1, distinctive synthetic expression may or may not align with identical synthetic masks. This process establishes aligned groupings between synthetic images $\{\mathbf{x}^\text{syn}_{\text{img}, i}\}_{i=1}^{m}$ and textual expressions $\{\mathbf{x}^\text{syn}_{\text{txt}, j}\}_{j=1}^{n}$ through semantic-aware mask consensus. This validation process generates refined segmentation masks $\mathbf{M}^\text{syn}_{(i,j)}$ for each semantically aligned group as shown in Fig.~\ref{fig:grouping}.

First, we generate $n$ pseudo segmentation masks $\hat{\mathbf{M}}^\text{psd}_{(i,j)}$ for each synthetic image $\mathbf{x}^\text{syn}_{\text{img}, i}$ using a pretrained RES model~\cite{lai2024lisa}, producing $m \times n$ candidate masks. An identification function $\mathbb{I}$ converts continuous mask values into binary masks through confidence thresholding:
\[
\mathbf{M}^\text{psd}_{(i,j)} = \mathbb{I}\left(\hat{\mathbf{M}}^\text{psd}_{(i,j)}\right) = \begin{cases} 
1 & \text{if } \hat{\mathbf{M}}^\text{psd}_{(i,j)} \geq 0.5 \\
0 & \text{otherwise}
\end{cases}.
\]
This per-image mask generation ensures high-quality pseudo segmentation masks construction as each synthetic image mostly contains only one primary object~\cite{ha2024finding} and at the same time, pretrained RES models are good at generate the masks of a single target.

Next, we compute pairwise Intersection-over-Union (IoU) scores between all binary mask pairs \(\mathbf{M}^\text{psd}_{(i, j_1)}\) and \(\mathbf{M}^\text{psd}_{(i, j_2)}\) derived from different textual expressions for the same image. The mean IoU (mIoU) across all images for each expression pair is:
\[
\text{mIoU}(j_1, j_2) = \frac{1}{m} \sum_{i=1}^{m} \text{IoU}\left(\mathbf{M}^\text{psd}_{(i, j_1)}, \mathbf{M}^\text{psd}_{(i, j_2)}\right).
\]

Expression pairs achieving mIoU scores exceeding threshold $\tau$ are clustered into consensus groups $G_k$, while singleton expressions below this threshold are discarded. We then compute refined masks through per-group averaging:
\[
\hat{\mathbf{M}}^\text{syn}_{(i, j)} = \frac{1}{|G_k|} \sum_{j \in G_k} \hat{\mathbf{M}}^\text{psd}_{(i, j)},
\]
where $G_k$ denotes the consensus group containing expression $\mathbf{x}^\text{syn}_{\text{txt}}$.

Finally, we apply the identification function to these refined masks:
\[
\mathbf{M}^\text{syn}_{(i, j)} = \mathbb{I}\left(\hat{\mathbf{M}}^\text{syn}_{(i, j)}\right),
\]
yielding binary masks where pixels with averaged values $>$ 0.5 are assigned 1, and 0 otherwise. For ease of notation, $\mathbf{M}^\text{syn}_{(i, j)}$ is denoted $\mathbf{M}^\text{syn}_{k}$ where k ranges from $1$ to $l$.

By proceeding these things, we generated synthetic triplets consisting of images, text expressions, and segmentation masks: $\{\mathbf{x}^\text{syn}_{\text{img}, i}\}^{m}_{i=1}$, $\{\mathbf{x}^\text{syn}_{\text{txt}, j}\}^{n}_{j=1}$, and $\{\mathbf{M}^\text{syn}_k\}^{l}_{k=1}$, respectively. These were derived from a single real image and its corresponding mask, denoted as $\mathbf{x}^\text{real}_{\text{img}}$ and $\mathbf{M}^\text{real}$  (see \supp~\ref{sec:step2_exam} for visualization).

\vspace{-1mm}
\subsection{Step 3: Domain-aware Augmentation for Images-Masks and Text}
\vspace{-1mm}

\nbf{Multi-Target aimed Image-mask Augmentation} In RES in the wild, it is important for a model to distinguish the same objects with different attributes. We apply mosaic augmentation~\cite{xie2024mosaicfusion, ha2024finding} by constructing composite arrangements with synthetic images and masks to facilitate multi-object/target segmentation as shown in Fig.~\ref{fig:augmentation}. In this process, each mosaic randomly adopts either a \(2 \times 2\) grid (1 real image \(\mathbf{x}^\mathrm{real}_\mathrm{img}\) + 3 synthetic \(\mathbf{x}^\mathrm{syn}_\mathrm{img,i}\)) or \(3 \times 3\) grid (1 real + 8 synthetic), with corresponding masks \(\mathbf{M}^\mathrm{real}\) and \(\mathbf{M}^\mathrm{syn}_i\). This approach facilitates the accommodation of multiple target objects within a single augmented scene. Simultaneously, integrating real and synthetic images creates composite visuals that effectively mitigate the domain gap between real-world and synthetically generated images.

\nbf{Debiased Text Augmentation} We observe that training data commonly demonstrates a significant bias toward specific object class terminology. This systematic bias results in segmentation failures when processing input queries containing alternative synonymous expressions. We apply the text augmentation that probabilistically replaces head and sub-nouns with their corresponding superclasses. For instance, the phrase “\textit{woman with hair tied up in a bun}” is transformed into “\textit{person with hair tied up in a bun}” with probability \(p\), thereby shifting the emphasis toward the action and mitigating gender bias (see \supp~\ref{sec:superclass} for superclass and original words). This approach is especially useful for our synthetic data, as implementing such augmentation with conventional RES datasets (e.g., RefCOCO) would present significant challenges: these collections typically contain multiple objects and provide only minimal discriminative cues necessary for precise target identification. Consequently, modifying expressions with superclass terminology could inadvertently introduce non-distinctive queries, potentially resulting in imprecise segmentation masks through the inclusion of non-target objects (\textcolor{magenta}{false negative masks} in Fig.~\ref{fig:augmentation}).

\vspace{-1mm}
\section{Experiments}
\begin{table*}[!t]
    \centering
    {
        \resizebox{\linewidth}{!}{
            \begin{tabular}{ l | c c | c c | c c | c c | c c }
                \toprule
                
                \multirow{3}*{Model} & \multicolumn{2}{c|}{\multirow{2}{*}{Training data}} & \multicolumn{2}{c|}{val} & \multicolumn{6}{c}{test} \\ 
                
                \specialrule{0em}{0pt}{1pt}
                \cline{4-11}
                \specialrule{0em}{0pt}{1pt}
                
                ~ & ~ & ~ & \multicolumn{2}{c|}{overall} & \multicolumn{2}{c|}{many attribute} & \multicolumn{2}{c|}{shared attribute} & \multicolumn{2}{c}{overall} \\
        
                \specialrule{0em}{0pt}{1pt}
                \cline{2-11}
                \specialrule{0em}{0pt}{1pt}
                
                ~ & real & synthetic & gIoU & cIoU & gIoU & cIoU & gIoU & cIoU & gIoU & cIoU \\ 
                
                \specialrule{0em}{0pt}{1pt}
                \hline
                \specialrule{0em}{0pt}{1pt}
                
                \multirow{3}{*}{LISA-7B~\cite{lai2024lisa}} & \cmark & \xmark & 37.1 & 40.3 & \textbf{30.1} & 25.0 & 39.3 & 36.0 & 34.5 & 32.5 \\
        
                ~ & \cmark & \cmark FreeMask~\cite{yang2023freemask} & 38.8 & 42.0 & 25.8 & 25.3 & 37.1 & 36.4 & 31.6 & 33.3 \\
        
                ~ & \cmark & \cmark SynRES (Ours) &
                \textbf{41.3}\textcolor{red}{\dt{+4.2}} &
                \textbf{46.1}\textcolor{red}{\dt{+5.8}} &
                29.4\textcolor{blue}{\dt{-0.7}} &
                \textbf{26.3}\textcolor{red}{\dt{+1.3}} &
                \textbf{43.6}\textcolor{red}{\dt{+4.3}} &
                \textbf{43.3}\textcolor{red}{\dt{+7.3}} &
                \textbf{36.5}\textcolor{red}{\dt{+2.0}} &
                \textbf{38.1}\textcolor{red}{\dt{+5.6}} \\
                
                \specialrule{0em}{0pt}{1pt}
                \hline
                \specialrule{0em}{0pt}{1pt}
                
                \multirow{2}{*}{LISA-13B~\cite{lai2024lisa}} & \cmark & \xmark & 44.0 & 45.1 & 35.9 & 31.6 & 36.9 & 37.8 & 37.7 & 36.0 \\
                
                ~ & \cmark & \cmark SynRES (Ours) &
                \textbf{47.7}\textcolor{red}{\dt{+3.7}} &
                \textbf{51.8}\textcolor{red}{\dt{+6.7}} &
                \textbf{36.5}\textcolor{red}{\dt{+0.6}} &
                \textbf{33.4}\textcolor{red}{\dt{+1.8}} &
                \textbf{44.7}\textcolor{red}{\dt{+7.8}} &
                \textbf{43.9}\textcolor{red}{\dt{+6.1}} &
                \textbf{40.5}\textcolor{red}{\dt{+2.8}} &
                \textbf{40.8}\textcolor{red}{\dt{+4.8}} \\
                
                \specialrule{0em}{0pt}{1pt}
                \hline
                \specialrule{0em}{0pt}{1pt}
                
                \multirow{2}{*}{GSVA-7B~\cite{xia2024gsva}} & \cmark & \xmark & 38.8 & 42.8 & \textbf{32.6} & 26.1 & 37.7 & 34.8 & 34.8 & 32.0 \\
                
                ~ & \cmark & \cmark SynRES (Ours) &
                \textbf{41.3}\textcolor{red}{\dt{+2.5}} &
                \textbf{46.5}\textcolor{red}{\dt{+3.7}} &
                32.4 \textcolor{blue}{\dt{-0.2}} &
                \textbf{26.3}\textcolor{red}{\dt{+0.2}} &
                \textbf{43.6}\textcolor{red}{\dt{+5.9}} &
                \textbf{41.5}\textcolor{red}{\dt{+6.7}} &
                \textbf{38.0}\textcolor{red}{\dt{+3.2}} &
                \textbf{36.9}\textcolor{red}{\dt{+4.9}} \\
                
                \bottomrule            
            \end{tabular}
        }
    }
    \vspace{-2mm}
    \caption{\wildres-ID segmentation results among \om{} (ours) and previous related works. Models trained only with real data use an additional 5000 fine-tuning steps on the validation set of \wildres-ID.}
    \label{table:refcocow_result}   
    \vspace{-3mm}
\end{table*}
\begin{table*}[!t]
    \centering
    \tabcolsep=0.25cm
    {
        \resizebox{0.95\linewidth}{!}{
            \begin{tabular}{ l | c c | c c | c c | c c | c c }
                \toprule
                
                \multirow{3}*{Model} & \multicolumn{2}{c|}{\multirow{2}{*}{Training data}} & \multicolumn{6}{c}{Domain source} & \multicolumn{2}{|c}{\multirow{2}{*}{average}}  \\ 
                
                \specialrule{0em}{0pt}{1pt}
                \cline{4-9}
                \specialrule{0em}{0pt}{1pt}
                
                ~ & ~ & ~ & \multicolumn{2}{c|}{CrowdHuman~\cite{shao2018crowdhuman}} & \multicolumn{2}{c|}{Cityscapes~\cite{Cordts2016Cityscapes}} & \multicolumn{2}{c|}{ARMBench~\cite{mitash2023armbench}} \\
    
                \specialrule{0em}{0pt}{1pt}
                \cline{2-11}
                \specialrule{0em}{0pt}{1pt}
                
                ~ & real & synthetic & gIoU & cIoU & gIoU & cIoU & gIoU & cIoU & gIoU & cIoU \\ 
                
                \specialrule{0em}{0pt}{1pt}
                \hline
                \specialrule{0em}{0pt}{1pt}
                
                \multirow{3}{*}{LISA-7B~\cite{lai2024lisa}} & \cmark & \xmark & \textbf{25.6} & 26.3 & 37.1 & 39.4 & 26.9 & 26.0  & 29.9 & 30.6 \\
    
                ~ & \cmark & \cmark FreeMask~\cite{yang2023freemask} & 23.5 & 26.0 & 34.4 & 36.9 & 24.5 & 26.1 & 27.5 & 29.7 \\
    
                ~ & \cmark & \cmark SynRES (Ours) & 24.8 & \textbf{27.3} & \textbf{40.5} & \textbf{43.2} & \textbf{35.8} & \textbf{33.3} & \textbf{33.7}\textcolor{red}{\dt{+3.8}} & \textbf{34.6}\textcolor{red}{\dt{+4.0}} \\
                
                \specialrule{0em}{0pt}{1pt}
                \hline
                \specialrule{0em}{0pt}{1pt}
                
                \multirow{2}{*}{LISA-13B~\cite{lai2024lisa}} & \cmark & \xmark & 26.6 & 29.4 & 34.1 & 34.8 & 28.3 & 26.9 & 29.7 & 30.4 \\
                
                ~ & \cmark & \cmark SynRES (Ours) & \textbf{28.8 }& \textbf{30.0} & \textbf{39.9} & \textbf{43.6} & \textbf{39.1} & \textbf{37.7} & \textbf{35.9}\textcolor{red}{\dt{+6.2}} & \textbf{37.1}\textcolor{red}{\dt{+6.7}} \\
                
                \specialrule{0em}{0pt}{1pt}
                \hline
                \specialrule{0em}{0pt}{1pt}
                
                \multirow{2}{*}{GSVA-7B~\cite{xia2024gsva}} & \cmark & \xmark & 19.8 & 24.6 & 35.9 & 36.5 & 25.1 & 23.6 & 26.9 & 28.2 \\
                
                ~ & \cmark & \cmark SynRES (Ours) & \textbf{21.4} & \textbf{25.5} & \textbf{41.2} & \textbf{44.7} & \textbf{33.6} & \textbf{32.6} & \textbf{32.1}\textcolor{red}{\dt{+5.2}} & \textbf{34.2}\textcolor{red}{\dt{+6.0}} \\
                
                \bottomrule            
            \end{tabular}
        }
    }
    \vspace{-2.5mm}
    \caption{\wildres-DS segmentation results among SynRES (ours) and previous related works. All models trained with additional 5000 fine-tuning steps on the validation set of \wildres-ID.} 
    \label{table:wildseg_result}   
\vspace{-3mm}
\end{table*}

\subsection{Experimental Setups}
\nbf{Network Architecture} To demonstrate the effectiveness of our approach, we applied our method to RES models using large multimodal models LISA-7B, 13B~\cite{lai2024lisa}, and GSVA-7B~\cite{xia2024gsva}. These implementations were selected because they provide fully open-sourced training code, datasets, and pre-trained weights. For other architectural components requiring hyperparameters (\eg, projection layer dimensions), we adhered to the configurations specified in the original papers.

\nbf{Implementation Details} We utilized 4 NVIDIA A6000 48G GPUs for training. Other settings, including the optimizer and learning rate scheduler, were consistent with those of the network architecture. Specifically, we generated $m = 6$ synthetic images per referring target and up to $n = 5$ synthetic expressions based on RefCOCO+. This resulted in a total of 84,480 synthetic images, 78,263 expressions, and 21,592 masks. The default mIoU threshold $\tau$ in \om{} step 2 was set to 0.65, and the replacement probability $p$ for expression data augmentation was set to 0.7.

\nbf{Evaluation Metrics} We employed two evaluation metrics: gIoU and cIoU. The gIoU metric represents the average IoU values calculated per image, whereas cIoU is computed as the cumulative intersection over the cumulative union. Given cIoU's high sensitivity to large-area objects and higher-resolution images, gIoU was primarily used for evaluation~\cite{lai2024lisa}.

\subsection{\wildres: RES in the Wild}
\begin{table*}[t]
    \centering
    \tabcolsep=0.22cm
    {
        \resizebox{\linewidth}{!}{
        \begin{tabular}{ l | c c | c c c | c c c | c c | c c c }
            \toprule
            
            \multirow{2}*{Model} & \multicolumn{2}{c|}{Training data} & \multicolumn{3}{c|}{RefCOCO} & \multicolumn{3}{c|}{RefCOCO+}  & \multicolumn{2}{c|}{RefCOCOg} & \multicolumn{3}{c}{gRefCOCO} \\ 
            
            \specialrule{0em}{0pt}{1pt}
            \cline{2-14}
            \specialrule{0em}{0pt}{1pt}
            
            ~ & real & synthetic & val & testA & testB & val & testA & testB & val(U) & test(U) & val & testA & testB \\ 
            
            \specialrule{0em}{0pt}{1pt}
            \hline
            \specialrule{0em}{0pt}{1pt}



            CRIS~\cite{wang2022cris} & \cmark & \xmark & 70.5 & 73.2 & 66.1 & 62.3 & 68.1 & 53.7 & 59.9 & 60.4 & 56.3 & 63.4 & 51.8 \\

            LAVT~\cite{yang2022lavt} & \cmark & \xmark & 72.7 & 75.8 & 68.8 & 62.1 & 68.4 & 55.1 & 61.2 & 62.1 & 58.4 & 65.9 & 55.8 \\
            
            ReLA~\cite{liu2023gres} & \cmark & \xmark & 73.8 & 76.5 & 70.2 & \textbf{66.0} & \textbf{71.0} & 57.7 & 65.0 & 66.0 & 63.6 & 70.0 & 61.0 \\
            
            X-Decoder~\cite{zou2023xdecoder} & \cmark & \xmark & - & - & - & - & - & - & 64.6 & - & - & - & - \\

            SEEM~\cite{zou2023seem} & \cmark & \xmark & - & - & - & - & - & - & 65.7 & - & - & - & - \\
            
            \specialrule{0em}{0pt}{1pt}
            \hline
            \specialrule{0em}{0pt}{1pt}
            

            \multirow{3}*{LISA-7B~\cite{lai2024lisa}} & \cmark & \xmark & 73.4 & 75.7 & 70.0 & 62.3 & 66.8 & 56.2 & 67.3 & 68.1 & - & - & - \\

             & \cmark & \cmark FreeMask~\cite{yang2023freemask} & 72.4 & 74.8 & 68.3 & 59.7 & 65.3 & 52.9 & 65.4 & 66.0 & - & - & - \\

             & \cmark & \cmark SynRES (Ours) & \textbf{75.5}\textcolor{red}{\dt{+2.1}} & \textbf{77.0}\textcolor{red}{\dt{+2.3}} & \textbf{72.0}\textcolor{red}{\dt{+2.0}} & \textbf{64.8}\textcolor{red}{\dt{+2.5}} & \textbf{69.2}\textcolor{red}{\dt{+2.4}} & \textbf{58.8}\textcolor{red}{\dt{+2.6}} & \textbf{68.5}\textcolor{red}{\dt{+1.2}} & \textbf{69.8}\textcolor{red}{\dt{+1.7}} & - & - & - \\

            \specialrule{0em}{0pt}{1pt}
            \hline
            \specialrule{0em}{0pt}{1pt}

            \multirow{2}*{GSVA-7B~\cite{xia2024gsva}} & \cmark & \xmark & \textbf{76.0} & \textbf{77.8} & \textbf{72.7} & 63.6 & \textbf{68.2} & 58.3 & 69.9 & 70.7 & 72.6 & 73.7 & 66.5 \\

             & \cmark & \cmark SynRES (Ours) & \textbf{76.0}\dt{+0.0} & 77.5\textcolor{blue}{\dt{-0.3}} & 72.6\textcolor{blue}{\dt{-0.1}} & \textbf{64.2}\textcolor{red}{\dt{+0.6}} & \textbf{68.2}\dt{+0.0} & \textbf{59.0}\textcolor{red}{\dt{+0.7}} & \textbf{70.1}\textcolor{red}{\dt{+0.2}} & \textbf{70.9}\textcolor{red}{\dt{+0.2}} & \textbf{73.6}\textcolor{red}{\dt{+1.0}} & \textbf{74.6}\textcolor{red}{\dt{+0.9}} & \textbf{67.9}\textcolor{red}{\dt{+1.4}} \\
            


            
            \bottomrule            
        \end{tabular}
        }
    }
    \vspace{-0.2cm}
    \caption{Classic referring segmentation results (gIoU) compared w/ and w/o \om{} on existing RES models. All the numbers from LISA and GSVA are reproduced using their official repository and weights.}
    \label{table:refcocos_result}
\vspace{-0.4cm}
\end{table*}

\nbf{Settings}
We fine-tune officially released pretrained weights augmented with \om{}, utilizing datasets originally designed for: semantic segmentation~\cite{zhou2017ade20k,caesar2018coco,ramanathan2023paco,chen2014pascalpart,he2022partimagenet}, RES~\cite{yu2016refcocorefcoco+,rohrbach2016refclef,mao2016refcocog}, VQA~\cite{liu2024vqa1,liu2024vqa2}, and ReasonSeg for LISA~\cite{lai2024lisa}. For GSVA fine-tuning~\cite{xia2024gsva}, we incorporate gRefCOCO~\cite{liu2023gres}. Training maintains a 9:3:3:1:4 data ratio across semantic segmentation, classic RES, VQA, ReasonSeg, and \om{}.

All experiments use official pretrained weights for each RES model, training for 5,000 steps with \wildres-ID validation every 100 steps regardless of synthetic data inclusion. Notably, the \wildres-DS remains \emph{unused} during validation.

\nbf{Results}
Tab.~\ref{table:refcocow_result} compares \wildres-ID performance between baseline models and \om-enhanced variants. The LISA-7B baseline achieves 37.1 gIoU and 40.3 cIoU on validation data, while its \ours-augmented version reaches 41.3 gIoU (+4.2) and 46.1 cIoU (+5.8). Similar improvements emerge in multi-target and multi-attribute test scenarios. This scaling trend persists with LISA-13B, where synthetic data yields comparable gains of +2.8 gIoU and +4.8 cIoU over the LISA-13B baseline. The pattern extends to GSVA architectures, demonstrating consistent cross-model improvements that validate synthetic augmentation efficacy.

Tab.~\ref{table:wildseg_result} demonstrates \wildres-DS experiments simulating domain shifts using the CrowdHuman, Cityscapes, and ARMBench datasets. The ARMBench subset shows remarkable improvements exceeding 10 gIoU points across all model sizes. Models incorporating the \om{} achieve superior segmentation accuracy in diverse domains, outperforming real-data only baselines.

\subsection{Classic RES Benchmarks}
\nbf{Settings}  
We leveraged official LISA weights that incorporated the ReasonSeg validation set during training. Performance was evaluated on the ReasonSeg test set for epoch-wise model selection. For GSVA, where official weights exclude the ReasonSeg validation set, we perform model selection using ReasonSeg validation set. Other configurations remain consistent with the previous section.


\nbf{Results}
In Tab.~\ref{table:refcocos_result}, we observed that using \om{} consistently achieved the best performance across the not only single-target RefCOCO series but also multi-target and no-target included gRefCOCO. Interestingly, incorporating FreeMask~\cite{yang2023freemask} led to a decline in performance. This outcome highlights that simply adding synthetic datasets does not guarantee improved performance in RES; instead, it underscores the importance of effectively handling such datasets to achieve positive results.

\begin{table}
\begin{center}
\setlength{\tabcolsep}{2mm}{
\renewcommand\arraystretch{0.9}
\resizebox{\linewidth}{!}{
\begin{tabular}{l|ccc|ccc}
\toprule
\multirow{2}{*}{\tabincell{l}{Model w/ \\ LISA-7B}} & \multicolumn{3}{c|}{Modifications} & \multicolumn{2}{c}{\wildres-ID val} \\
&
{Step 2.}
&
{Mosaic Aug.}
& 
{Text Aug.}
& gIoU & cIoU \\
\midrule
\om{}                   & \cmark & \cmark & \cmark & 41.3 & 46.1 \\
                        & \cmark & \cmark & \xmark & 40.1 & 44.0 \\
                        & \xmark & \cmark & \cmark & 39.1 & 41.9 \\
                        & \cmark & \xmark & \xmark & 35.5 & 38.3 \\
\midrule
Only Real               & \xmark & \xmark & \xmark & 37.1 & 40.3 \\

\bottomrule
\end{tabular}}
}
\end{center}
\vspace{-0.23in}
\caption{Ablation study on the core designs of \om{}. \cmark means the employment of the component while \xmark means not.}
\label{tab:abl}
\vspace{-0.5cm}
\end{table}
\subsection{Qualitative comparison}
\vspace{-2pt}
\begin{figure*}[tp]
    \centering
    \vspace{-2mm}
    \includegraphics[clip, width=0.94\textwidth]{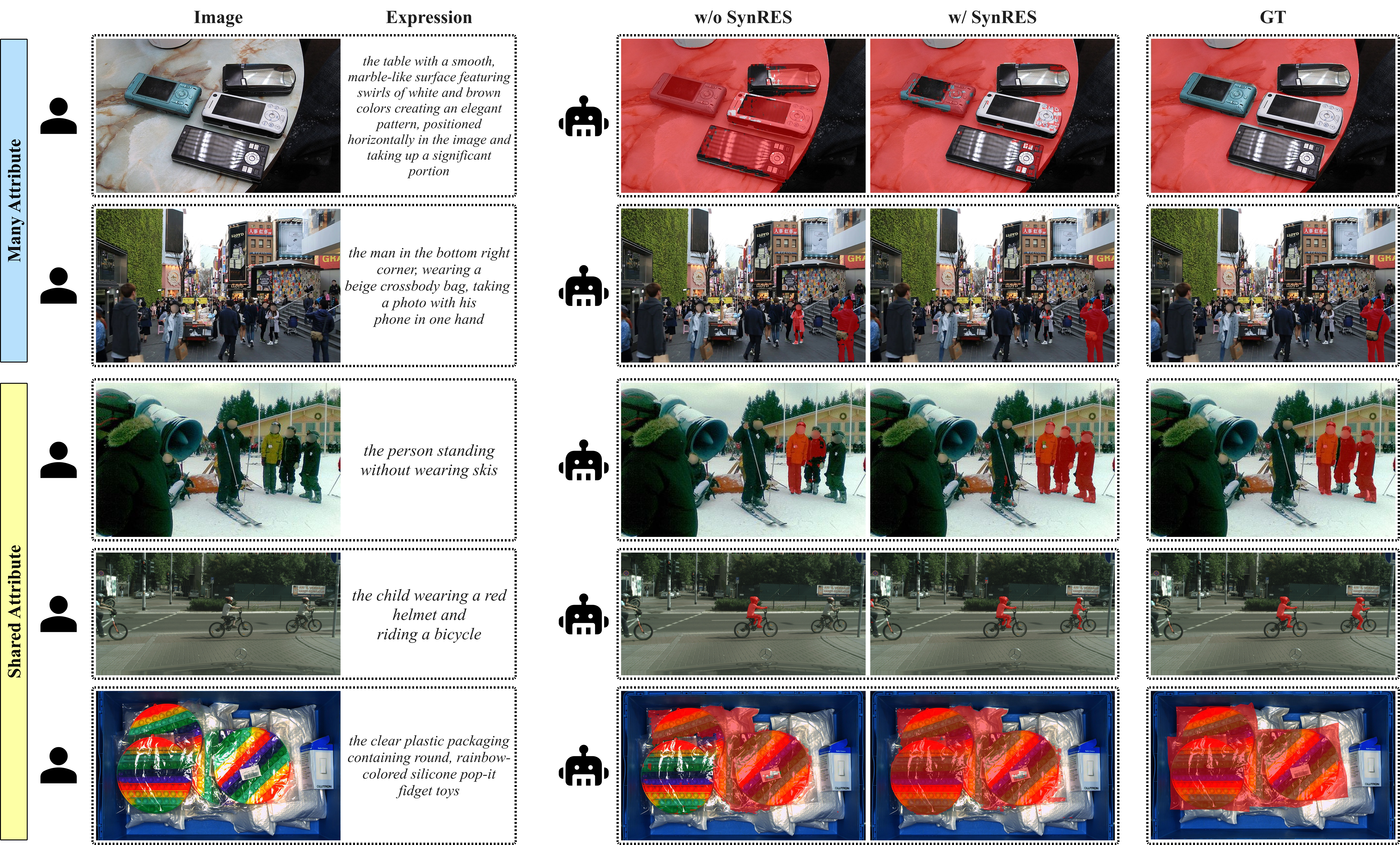}
    \vspace{-2mm}
    \caption{Qualitative results on \wildres, comparing LISA trained w/o and w/ \om{}.} 
    \label{fig:result_vis}
    \vspace{-5mm}
\end{figure*}

We compared the output masks of the LISA fine-tuned with \om{} and the LISA without \om{} on the \wildres dataset, as shown in Fig.~\ref{fig:result_vis}. The visualization shows that fine-tuning the model with \om{} reduces false negatives by capturing ground truth objects in multi-target scenarios involving shared attributes. Additionally, it prevents false positives by accurately understanding the meaning of each words in cases with many attribute expressions.

\vspace{-2pt}
\subsection{Ablation Study}
\vspace{-2pt}

Tab.~\ref{tab:abl} presents an ablation study evaluating each component in \om{}. The results show a gradual decline in performance, with the gIoU decreasing from 41.3 to 40.8 and 35.5 when text augmentation and mosaic augmentation were excluded, respectively. Notably, when both vision and text augmentations were omitted, the performance fell below the baseline value of 37.1, which was obtained without training on the synthetic dataset. This performance degradation can be attributed to two primary factors: first, the absence of vision augmentation restricts the model to the Wild RES task, which lacks multiple targets, leading to poor outcomes in the Shared Attribute portion; second, a domain gap between synthetic and real images further degrades performance. Moreover, the performance improvement with text augmentation confirms superclass replacement's effectiveness for contextual expression understanding in complex scenes by focusing on the description of the target. These findings underscore the importance of both text and vision augmentation in enhancing the effectiveness of the \om{} pipeline for complex visual grounding tasks. Additionally, adopting only domain-aware augmentation without mask refinement in \om{} step 2 shows limited performance, proving the necessity of refining raw pseudo masks.

\vspace{-1mm}

\vspace{-2pt}
\section{Conclusion}
\vspace{-2pt}
\label{sec:conclusion}

In this study, we introduce a new RES task aimed at addressing more complex reasoning in the real-world, dynamic scenarios, \wildres. Specifically, we focus on cases where expressions refer to multiple targets with shared attributions settings or require an increased number of attributes to identify a single target within such environments. To benchmark this task, we not only developed \wildres-ID within the same domain as existing RES datasets like RefCOCO but also extended its applicability to autonomous driving and robotics domains through the \wildres-DS. Additionally, we propose a method to enhance the performance of existing models in \wildres by leveraging densely paired synthetic image-expression-mask triplet generation and augmentation, \om{}. We hope that our work broadens the scope of language-based segmentation and contributes to advancements in this field.


{\small
\bibliographystyle{ieeenat_fullname}
\bibliography{11_references}
}

\clearpage \appendix \renewcommand{\thesection}{\Alph{section}}
\renewcommand{\thefigure}{\Alph{figure}}
\renewcommand{\thetable}{\Alph{table}}
\setcounter{page}{1}
\setcounter{section}{0}
\setcounter{figure}{0}
\setcounter{table}{0}

\startcontents
\printcontents{l}{1}{\section*{Supplementary Materials}}
\FloatBarrier

\section{Details of \wildres}
\label{sec:appendix_wildres_info} 

\begin{table}[t]
    \centering
    \resizebox{\columnwidth}{!}{%
        \begin{tabular}{c c c c c c}
            \toprule
            Type & Domain & Split & Attribute & Image & Expression \\
            \midrule
            \multirow{6}{*}{Wildseg-ID} & \multirow{6}{*}{MSCOCO} & \multirow{3}{*}{Validation} & Many Attribute & 100 & 138 \\
            & & & Shared Attribute & 104 & 127 \\ 
            \cmidrule(lr){4-6}
            & & & Total & 196 & 265 \\ 
            \cmidrule(lr){3-6}
            & & \multirow{3}{*}{Test} & Many Attribute & 108 & 133 \\
            & & & Shared Attribute & 115 & 124 \\
            \cmidrule(lr){4-6}
            & & & Total & 215 & 257 \\
            \midrule
            \multirow{3}{*}{Wildseg-DS} & CrowdHuman & \multirow{3}{*}{Test} & Many Attribute & 101 & 212 \\
            & Cityscapes & & Shared Attribute & 105 & 120 \\
            & Armbench & & Shared Attribute & 107 & 120 \\
            \midrule
            Total & & & & 724 & 974 \\
            \bottomrule
        \end{tabular}%
    }
    \caption{Detailed Numbers of WildRES}
    \label{tab:wildres_info}
\end{table}

Tab.~\ref{tab:wildres_info} provides a detailed number of the WildRES dataset, which consists of the Wildseg-ID and Wildseg-DS subsets. The dataset is categorized based on domain, data split (validation/test), and attribute type (Many Attribute or Shared Attribute). 
In total, WildRES comprises 724 images and 974 expressions across different domains and attribute types. 

\section{Referring Expression Attributions}
\label{sec:attr_type}
\begin{table}[t]
    \renewcommand\arraystretch{1.2}
    \centering
    \footnotesize
    \setlength{\tabcolsep}{0.1mm}{
    \begin{tabular}{c>{\hspace{3mm}}c<{\hspace{3mm}}c}
    \specialrule{.1em}{.05em}{.05em} 
        \textbf{ID} & \textbf{Attribute} &  \textbf{Word}\\
        \cline{1-3}
        A1 & head noun & \textit{cat} \\
        A2 & sub noun & {\textit{bench, boat}} \\
        A3 & {color} & \textit{{green}} \\
        A4 & {size} & {\textit{big}} \\
        A5 & {absolute location relation} & {\textit{the center}} \\
        A6 & {relative location relation} & {\textit{on, next to, in}} \\
        A7 & {action} & {\textit{sitting}} \\
        A8 & {generic attribute} & {\textit{wooden}} \\
    \specialrule{.1em}{.05em}{.05em} 
    \end{tabular}}%
    \caption{Types of attributes and their corresponding words in the referring expression, ``\textit{the cat sitting on the bench next to big green wooden boat in the center of the image}''}
    \label{tab:refer_attr}%
\end{table}%

Based on~\cite{yu2024crec}, we refine eight expression attributes identified in existing datasets: \textit{head noun}, \textit{sub noun}, \textit{color}, \textit{size}, \textit{absolute location relation}, \textit{relative location relation}, \textit{action}, and \textit{generic attribute}. A detailed example is shown in Tab.~\ref{tab:refer_attr}. Using these attributes, we conduct quantitative analyses comparing the attribute distribution patterns between classic datasets and \wildres. This counting process is automated using ChatGPT-4o-mini~\cite{achiam2023gpt4o}.

\section{Differences with \wildres and Previous datasets}
\label{sec:wildres_diff}
\subsection{Differences from Previous Multi-Target RES Datasets}
Existing multi-target RES datasets like gRefCOCO~\cite{liu2023gres} and Ref-ZOM~\cite{hu2023beyond} predominantly rely on enumeration, conjunctions, and plural head nouns. These datasets frequently employ numerical references (``Three persons playing baseball'') that assume known referent quantities—an assumption often disconnected from real-world scenarios where shared attributes drive identification.

While gRefCOCO incorporates attribute-based expressions (``A except B'' or ``A that has B''), these constructions appear infrequently—only 39 and 78 instances in 259,859 expressions—confirming enumeration dominates. Similarly, Ref-ZOM employs template-based generation that combines expressions from one-to-one datasets or embeds category information into predefined structures, sacrificing natural language flexibility for consistency.

\wildres adopts a fundamentally different approach by eliminating explicit numerical references and plural head nouns. Our dataset relies exclusively on shared attributes with singular head nouns to indicate multiple objects, creating more context-driven references. We focus on identifying subsets within categories based on distinctive attributes rather than comprehensive enumeration. For example, ``person holding a camera'' may coexist with ``person not holding a camera'' instances, enabling precise differentiation without explicitly counting targets. This design better reflects natural language patterns where speakers rarely enumerate objects explicitly. Unlike gRefCOCO, \wildres excludes no-target expressions entirely.

\begin{figure}[t]
    \centering
    \includegraphics[width=\linewidth]{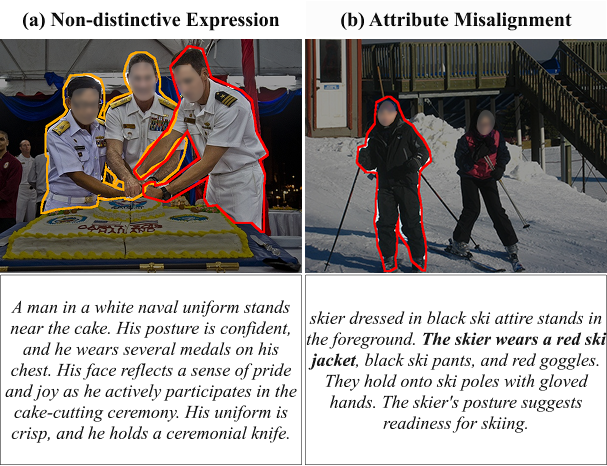}
    \caption{Pix2Cap misaligned examples}
    \label{fig:pix2cap_misaligned}
\end{figure}

\subsection{Difference from Pix2Cap}
Pix2Cap~\cite{you2025pix2cap} provides longer and more detailed captions from GPT-4V~\cite{hurst2024gpt4v} compared to traditional datasets. However, many captions fail to correspond to their respective objects accurately.

As illustrated in Fig.~\ref{fig:pix2cap_misaligned}, misalignment issues in the Pix2Cap dataset can be categorized into non-distinctive expressions and attribute mismatches. Non-distinctive expressions due to multiple objects (Fig.~\ref{fig:pix2cap_misaligned} (a)) occur when a referring expression corresponds to multiple objects, making the expression non-distinctive. The red-edged mask represents the Pix2Cap annotation, but additional objects matching the same description (highlighted in orange) result in incomplete coverage. Attribute misalignment (Fig.~\ref{fig:pix2cap_misaligned} (b)) arises when certain attributes within a referring expression do not accurately match the corresponding mask. The red mask represents the Pix2Cap annotation for the given expression, but the skier is not wearing a red ski jacket, leading to attribute misalignment.

To address these issues, we refined Pix2Cap-generated expressions or created new ones for images containing three or more objects, constructing an improved dataset with enhanced caption-to-object alignment while preserving Pix2Cap’s rich linguistic diversity.

\section{Structured Prompts for T2I Model}
\label{sec:t2i_prompt}

We randomly select one of two prompt structures as input for the T2I model in Sec.~\ref{sec:step1}:

\begin{tcolorbox}[colframe=black, colback=gray!10, rounded corners, width=\linewidth]
\small
\begin{itemize}
    \item \textit{photo of [aggregated description], hyper-realistic, 4k, realism, highly detailed, natural realistic background}
    \item \textit{cinematic scene [aggregated description], hyper-realistic, 4k, realism, highly detailed, natural realistic background}
\end{itemize}
\end{tcolorbox}
\noindent
These prompts bridge the synthetic-real domain gap while avoiding stylistic elements like cartoon or pixel art

\section{Superclass Replace Words}
\label{sec:superclass}
\begin{table}[t]
\centering
\tablestyle{3pt}{1.2}
\begin{tabular}{c|p{2cm}|p{5cm}}
\shline
Idx & Superclass Word & Original Word \\
\hline
1 & child & boy, girl, son, daughter \\
\hline
2 & kid & boy, girl, son, daughter \\
\hline
3 & adult & woman, women, man, men, female, male \\
\hline
4 & person & woman, women, man, men, female, male, boy, girl, guy \\
\hline
5 & their & his, her \\
\hline
6 & vehicle & car, bus, plane, train, airplane, truck, boat, motorcycle \\
\hline
7 & animal & bird, cow, bull, rabbit, bunny, dog, puppy, cat, zebra, elephant, horse, giraffe \\
\hline
8 & fruit & apple, banana \\
\hline
9 & vegetable & broccoli, carrot, cabbage, radish \\
\hline
10 & food & sandwich, hot dog, pizza, donut, doughnut, cake, hamburger \\
\hline
11 & electronic & tv, television, laptop, computer, keyboard, cell phone, smartphone \\
\hline
12 & furniture & chair, couch, sofa, bed, desk \\

\bottomrule
\end{tabular}
\caption{\textbf{Superclass and original words for text augmentation}}

\label{tab:superclass}
\end{table}

Tab.~\ref{tab:superclass} lists superclass replacements for text augmentation. Based on MSCOCO~\cite{caesar2018coco} classes and our superclass taxonomy, we select frequently occurring terms from \om{}'s synthetic expressions. To maintain expression consistency, gender-specific replacements trigger corresponding pronoun updates. For example, replacing ``\textit{boy}'' with ``\textit{child}'' in ``\textit{The boy holding his bag}'' simultaneously changes ``\textit{his}'' to ``\textit{their}'', resulting in ``\textit{The child holding their bag}''.

\section{Additional Experiments}
\subsection{Additional Baseline}

\begin{table}[t]
    \centering
    \resizebox{\columnwidth}{!}{%
    \renewcommand{\arraystretch}{0.8}
    \begin{tabular}{l *{10}{p{0.7cm}}}
        \toprule
        \multirow{2}{*}{Model} & \multicolumn{4}{c}{\wildres-ID} & \multicolumn{6}{c}{\wildres-DS} \\
        \cmidrule(lr){2-5} \cmidrule(lr){6-11}
        & \multicolumn{2}{c}{val} & \multicolumn{2}{c}{test} 
          & \multicolumn{2}{c}{crowdhuman} & \multicolumn{2}{c}{cityscapes} & \multicolumn{2}{c}{Armbench} \\
        \cmidrule(lr){2-3} \cmidrule(lr){4-5} \cmidrule(lr){6-7} \cmidrule(lr){8-9} \cmidrule(lr){10-11}
        & gIoU & cIoU & gIoU & cIoU & gIoU & cIoU & gIoU & cIoU & gIoU & cIoU \\
        \midrule
        GLaMM           & 39.4 & 42.2 & 36.6 & 34.8 & 33.9 & 31.5 & 19.5 & 23.7 & 32.3 & 29.2 \\
        GLaMM+\om{}    & \textbf{42.0} & \textbf{44.6} & \textbf{39.6} & \textbf{37.5} & \textbf{35.5} & \textbf{34.3} & \textbf{28.7} & \textbf{30.5} & \textbf{36.4} & \textbf{33.8} \\
        \bottomrule
    \end{tabular}%
    }
    \caption{GLaMM~\cite{rasheed2024glamm} results on \wildres dataset.}
    \label{table:glamm_result}
\end{table}

GLaMM~\cite{rasheed2024glamm} developed its own automated real dataset annotation pipeline and dataset (GranD) to address tasks such as Grounded Conversation Generation. We investigated whether this GLaMM can effectively perform on our benchmark \wildres{} and whether \om{} can further advance its capabilities.

\nbf{Settings}
We fine-tune officially released pretrained weights augmented with \om{}, utilizing datasets originally designed for RES~\cite{yu2016refcocorefcoco+,rohrbach2016refclef,mao2016refcocog}. Training maintains a 3:4 data ratio between classic RES and \om{}.

All experiments use official pretrained weights for each RES model, training for 5,000 steps with \wildres-ID validation every 100 steps regardless of synthetic data inclusion. Notably, the \wildres-DS remains \emph{unused} during validation.

\nbf{Results}
Tab.~\ref{table:glamm_result} demonstrates the effectiveness of synthetic data augmentation across both \wildres-ID and \wildres-DS benchmarks. Adding \om{} consistently outperforms the GLaMM, achieving improvements of +2.6 gIoU and +2.4 cIoU on validation data (42.0 vs 39.4 gIoU, 44.6 vs 42.2 cIoU) and +3.0 gIoU and +2.7 cIoU on test data. The enhancement extends across diverse domains in \wildres-DS, with performance gains ranging from +1.6 gIoU on CrowdHuman to +9.2 gIoU on Cityscapes, while Armbench shows substantial improvements of +4.1 gIoU and +4.6 cIoU. These results show that additional real annotated dataset (GranD) are not sufficient to address our complex RES scenarios. 

\subsection{Additional Benchmarks}
\begin{table}[ht]
\centering
\tiny
\setlength{\tabcolsep}{3pt}  
\renewcommand{\arraystretch}{0.9}
\begin{minipage}{0.33\columnwidth}
  \centering
  \begin{tabular}{lcc}
    \toprule
    \multicolumn{3}{c}{Ref-ZOM test} \\
    \midrule
     & gIoU  & cIoU  \\
     \cmidrule(lr){2-3}
    GSVA-7B        & 72.58 & 64.40 \\
    GSVA-7B+SynRES & \textbf{73.55} & \textbf{64.65} \\
    \bottomrule
  \end{tabular}
\end{minipage}%
\hfill
\begin{minipage}{0.6\columnwidth}
  \centering
  \begin{tabular}{l@{\hskip 4pt}cc@{\hskip 4pt}cc@{\hskip 4pt}cc}
    \toprule
    \multirow{2}{*}{Model} 
      & \multicolumn{2}{c}{\scriptsize LISA-7B} 
      & \multicolumn{2}{c}{\scriptsize LISA-13B} 
      & \multicolumn{2}{c}{\scriptsize GSVA-7B} \\
    \cmidrule(lr){2-3}\cmidrule(lr){4-5}\cmidrule(lr){6-7}
      & base & +\om{} & base & +\om{} & base & +\om{} \\
    \midrule
    val  & –   & –    & –    & –    & 45.13 & \textbf{45.78} \\
    test & \textbf{44.66} & 44.13 & 47.62 & \textbf{48.02} & 40.19 & \textbf{40.72} \\
    \bottomrule
  \end{tabular}
\end{minipage}
\caption{Results on Ref-ZOM test set (left) and ReasonSeg (right).}
\label{tab:additional_benchmark}
\end{table}

We further conducted experiments on Ref-ZOM~\cite{hu2023beyond} (multi-target datasets, Tab.~\ref{tab:additional_benchmark} left) and ReasonSeg~\cite{lai2024lisa} (long and implicit expressions, Tab.~\ref{tab:additional_benchmark} right). Results demonstrate that \om{} maintains or improves baseline performance across diverse scenarios. On Ref-ZOM, GSVA-7B+\om{} preserves strong performance with slight gains of +0.97 gIoU (73.55 vs 72.58). For ReasonSeg, the approach shows consistent stability across model scales: LISA-13B achieves +0.40 gIoU improvement (48.02 vs 47.62), while GSVA-7B maintains competitive performance with +0.65 gIoU on validation and +0.53 gIoU on test data.

\section{Discussions}
\begin{figure*}[t]
    \centering
    \includegraphics[width=0.7\linewidth]{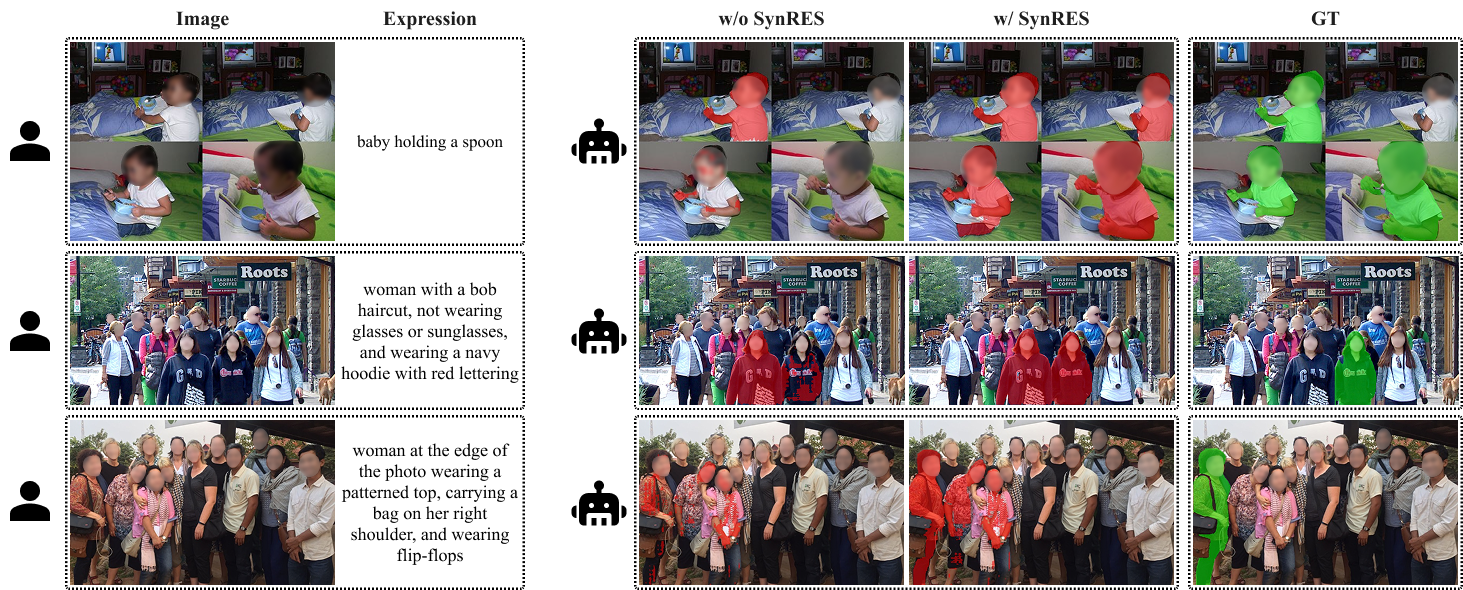}
    \caption{Failure Cases}
    \label{fig:limitations_failure}
\end{figure*}

\subsection{Hyperparameter Choices}
\begin{table}[h]
  \centering
  \makebox[\columnwidth][c]{%
    \begin{minipage}{0.2\columnwidth}
      \centering
      \small 
      \resizebox{\columnwidth}{!}{%
        \begin{tabular}{cc}
          \toprule
          $p$ & gIoU\\
          \midrule
          0.5 & 40.8 \\
          0.6 & 40.3 \\
          0.7 & \textbf{41.3} \\
          0.8 & 41.0 \\
          0.9 & 40.4 \\
          \bottomrule
        \end{tabular}%
      }
    \end{minipage}
    \hspace{1mm} 
    \begin{minipage}{0.2\columnwidth}
      \renewcommand*{\arraystretch}{1.1}
      \centering
      \small 
      \resizebox{\columnwidth}{!}{%
        \begin{tabular}{cc}
          \toprule
          $\tau$ & gIoU \\
          \midrule
          0.55 & 40.5 \\
          0.6  & 39.6 \\
          0.65 & \textbf{41.3} \\
          0.7  & 40.2 \\
          0.75 & 40.8 \\
          \bottomrule
        \end{tabular}%
      }
    \end{minipage}%
  }
  \caption{Hyperparameter ablation study results with replace probability $p$ and mIoU Threshold $\tau$.}
  \label{tab:ablation_hparam}
\end{table}
Tab.~\ref{tab:ablation_hparam} presents the results of experiments conducted by varying the replacement probability $p$, a hyperparameter of \om{}, from 0.4 to 0.9. Results demonstrate that the validation set of \wildres-ID consistently achieves gIoU scores above 40 across all configurations, indicating minimal performance sensitivity to $p$ variations. Similarly, when testing the mIoU threshold $\tau$ within 0.55-0.75, the minimum observed gIoU remains above 39.6, showing comparable robustness to $\tau$ selection.


\subsection{Performance Relative to Synthetic Data Quantity}
\begin{table}[t]
    \centering
    \resizebox{0.35\linewidth}{!}{
        \begin{tabular}{cc}
            \toprule
            Syn. Images (\%) & gIoU \\
            \midrule
            25\% & 40.7 \\
            50\%  & 40.8 \\
            75\% & 40.9 \\
            100\%  & \textbf{41.3} \\
            \bottomrule
        \end{tabular}%
    }
    \caption{Performance variations of LISA with \om{} on \wildres-ID validation set across different quantities of synthetic images.}
    \label{tab:num_syn_data}
\end{table}

We evaluate LISA's performance using progressively increasing quantities of synthetic training images (25\%, 50\%, 75\%, and 100\% of full dataset size). As shown in Tab.~\ref{tab:num_syn_data}, the gIoU metric demonstrates remarkable stability across data scales, with only 0.6 separating the 25\% and 100\% conditions. Notably, the 25\% configuration achieves 98.5\% of the maximum observed performance, suggesting efficient data utilization.


\begin{figure*}[tp]
    \centering
    \vspace{-2mm}
    \includegraphics[clip, width=0.95\textwidth]{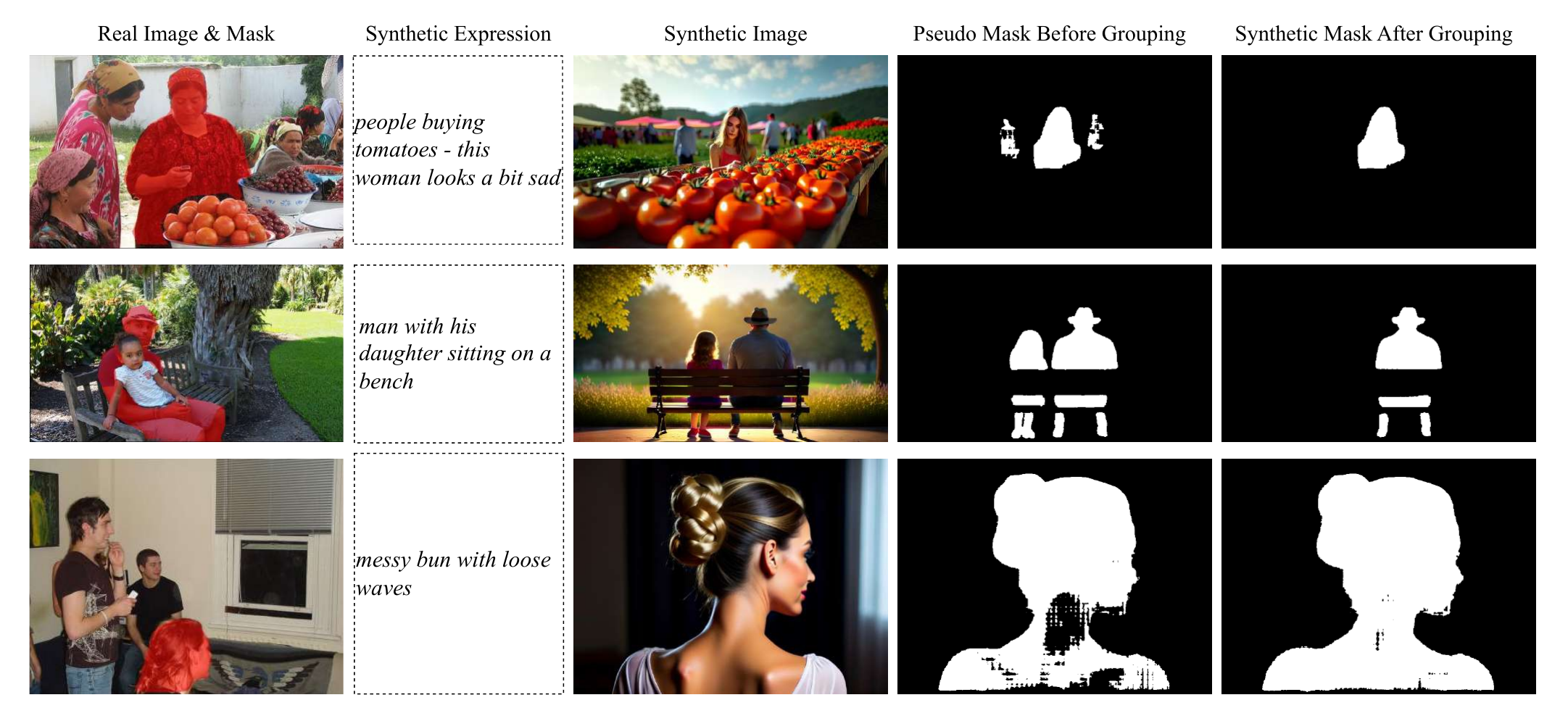}
    \vspace{-1mm}
    \caption{Examples of before and after of \om{} step 2.} 
    \label{fig:step2_exam}
    \vspace{-4mm}
\end{figure*}
\begin{figure*}[tp]
    \centering
    \vspace{-2mm}
    \includegraphics[clip, width=0.95\textwidth]{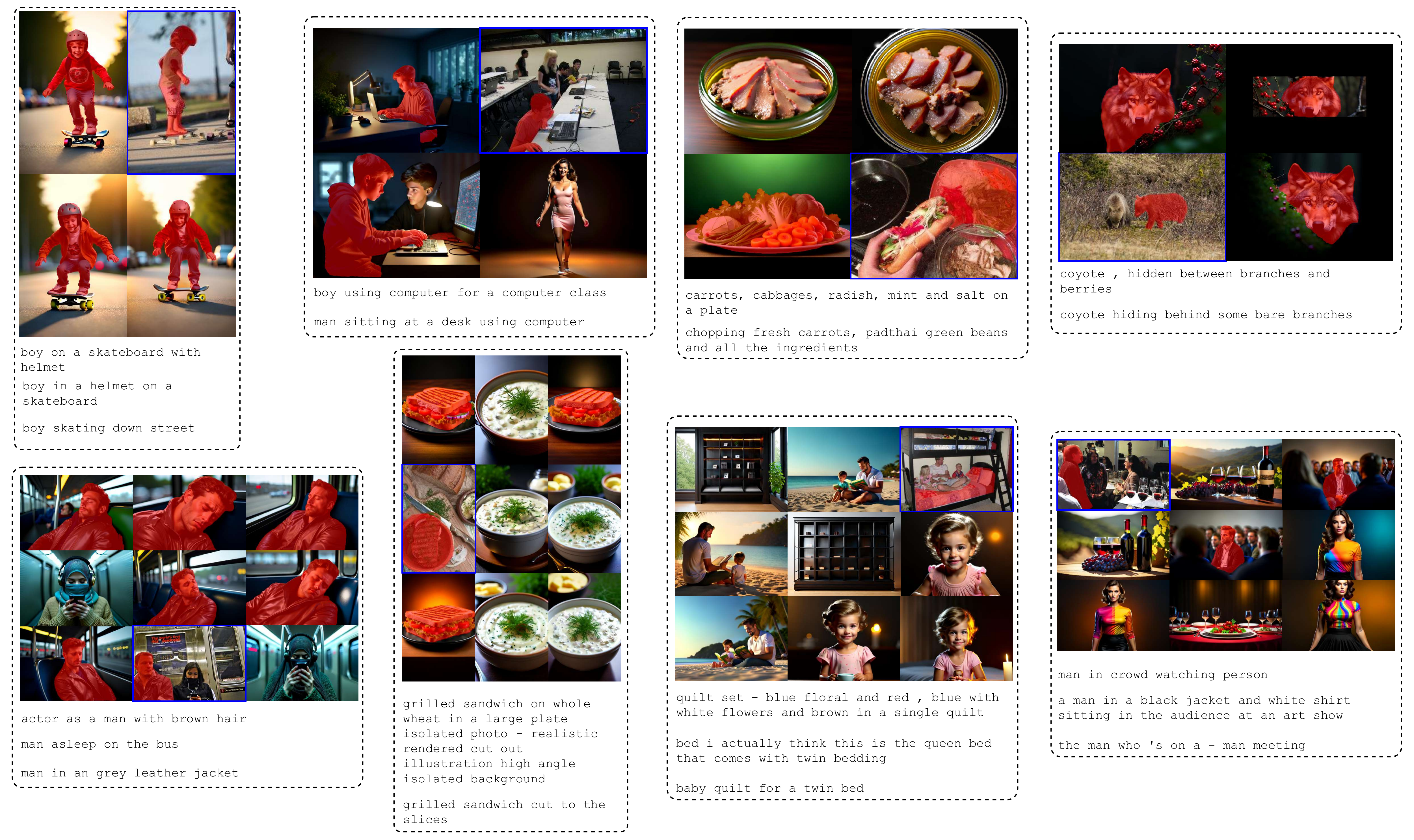}
    \vspace{-1mm}
    \caption{Examples of \om{} generated training data. Blue border is real image and mask.} 
    \label{fig:synres_exam}
    \vspace{-4mm}
\end{figure*}

\subsection{Computational Cost}
The process of generating the synthetic dataset does not incur additional training costs, as it leverages off-the-shelf models. However, the generation process does require time: generating synthetic expressions took approximately 20 hours, creating synthetic images took around 23 hours, and performing step 2 grouping required about 30 hours, all using four A6000 GPUs. Once the synthetic dataset is generated, effective results can be achieved with only 5,000 steps of additional fine-tuning—just $1/10$ of the training from scratch.

\subsection{Limitations and Future work}
Fig.~\ref{fig:limitations_failure} demonstrates characteristic failure cases involving over-segmentation beyond ground truth boundaries. These errors typically occur when visually similar objects appear immediately adjacent to targets. Our current mosaic augmentation struggles with such proximate object arrangements and limits composite images to 9 components, restricting training exposure to higher object densities. Implementing targeted copy-and-paste augmentation~\cite{fan2024divergen} that preserves original attributes could mitigate these limitations in future work.


\section{More Visualization}
\subsection{Refined Synthetic Mask Example Before and After \om{} Step 2}
\label{sec:step2_exam}
Fig.~\ref{fig:step2_exam} show mask refinement results, comparing synthetic segmentation outputs before and after \om{} in step 2.

\subsection{Additional \om{} Examples}
Fig.~\ref{fig:synres_exam} presents \om{} examples demonstrating synthetic dataset generation. Superclass replacement augmentation applied to these generated expressions following Tab.~\ref{sec:superclass}.

\subsection{Extended Qualitative Results}
We present additional visualization results in Fig.~\ref{fig:supplementary_visualization} to further demonstrate the effectiveness of the \wildres task with \om{}.

\begin{figure*}[t]
    \centering

    \begin{subfigure}{\textwidth}
        \includegraphics[width=\textwidth]{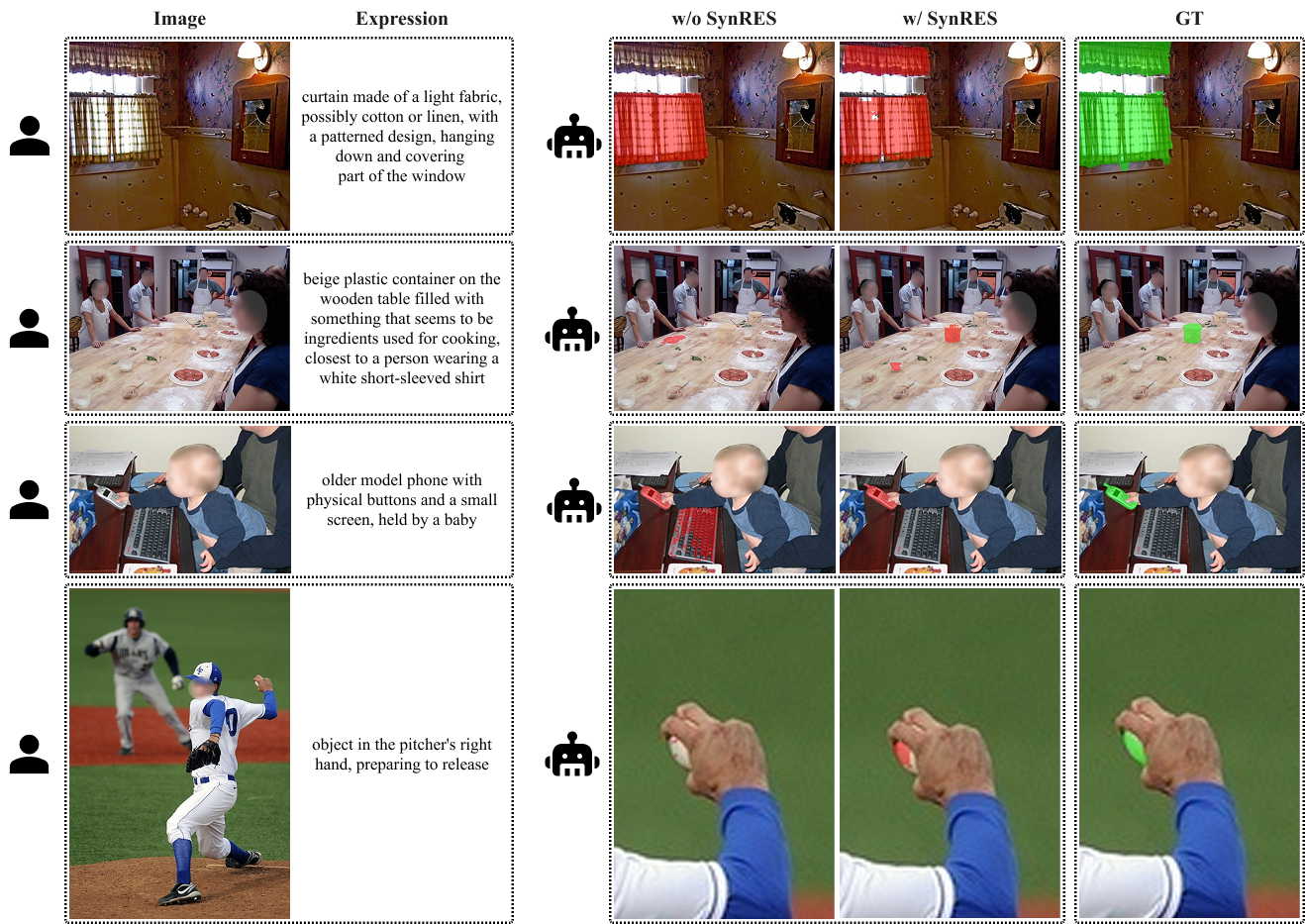}
        \caption{Visualization of Many Attribute in WildRES-ID}
        \label{fig:supplementary_ma}
    \end{subfigure}

    \begin{subfigure}{\textwidth}
        \includegraphics[width=\textwidth]{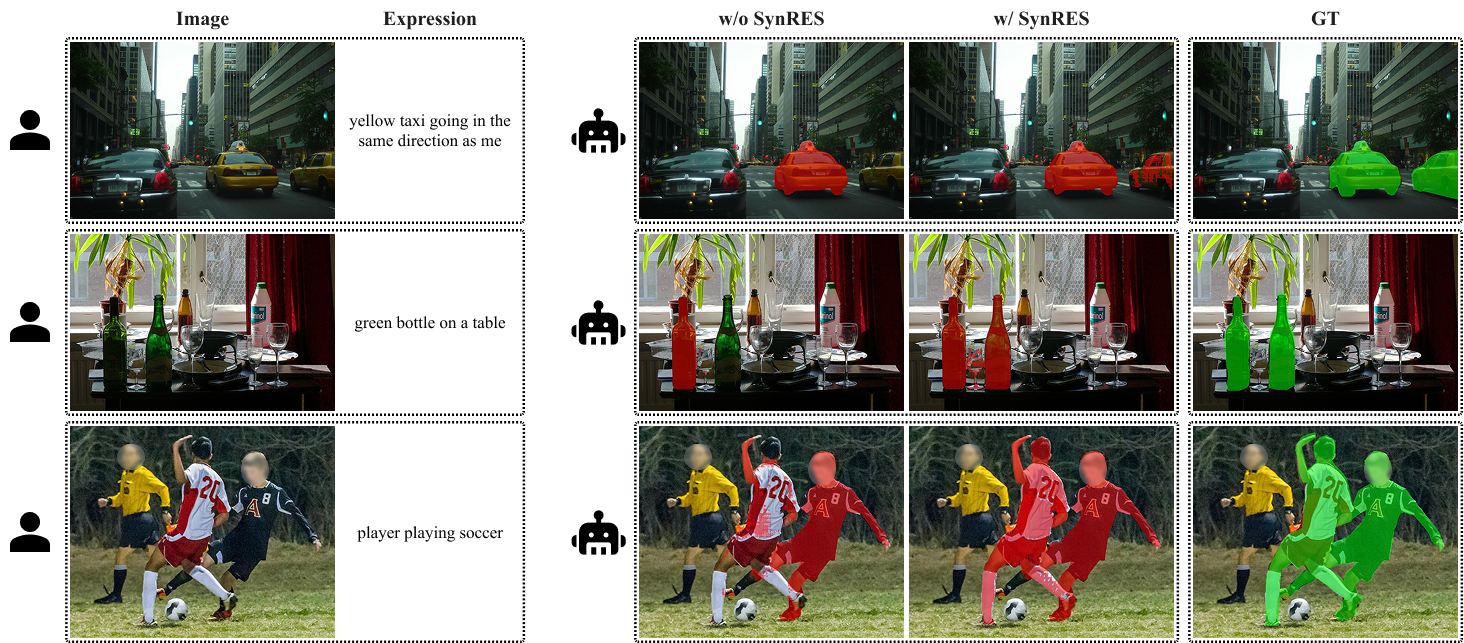}
        \caption{Visualization of Shared Attribute in WildRES-ID}
        \label{fig:supplementary_sa}
    \end{subfigure}
\end{figure*}

\begin{figure*}[t]\ContinuedFloat
    \begin{subfigure}{\textwidth}
        \includegraphics[width=\textwidth]{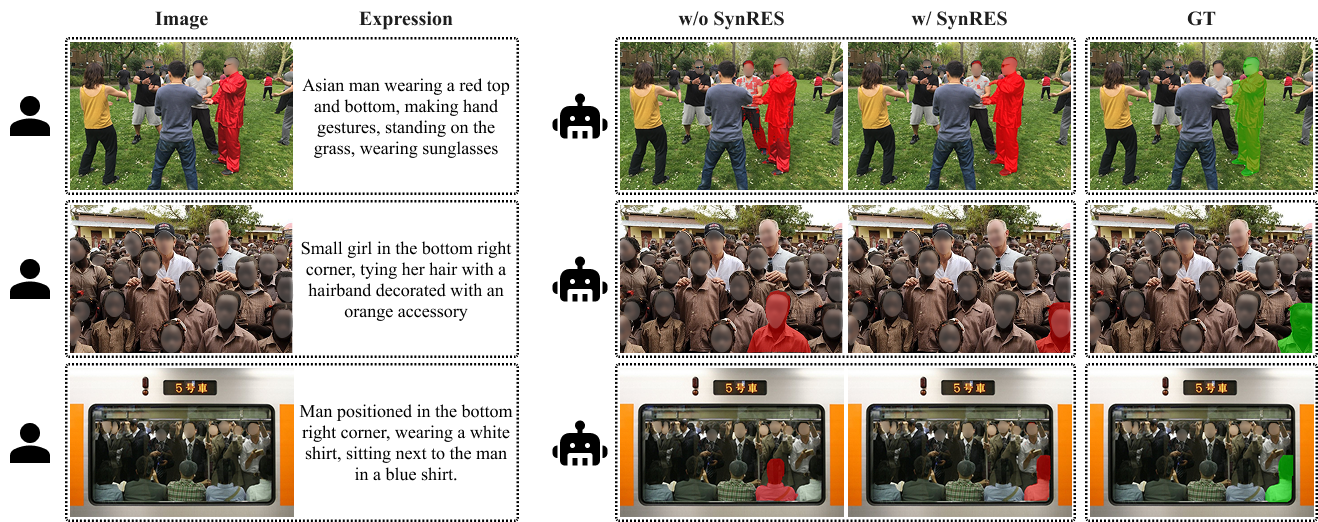}
        \caption{Visualization of CrowdHuman in WildRES-DS }
        \label{fig:supplementary_crowdhuamn}
    \end{subfigure}

    \begin{subfigure}{\textwidth}
        \includegraphics[width=\textwidth]{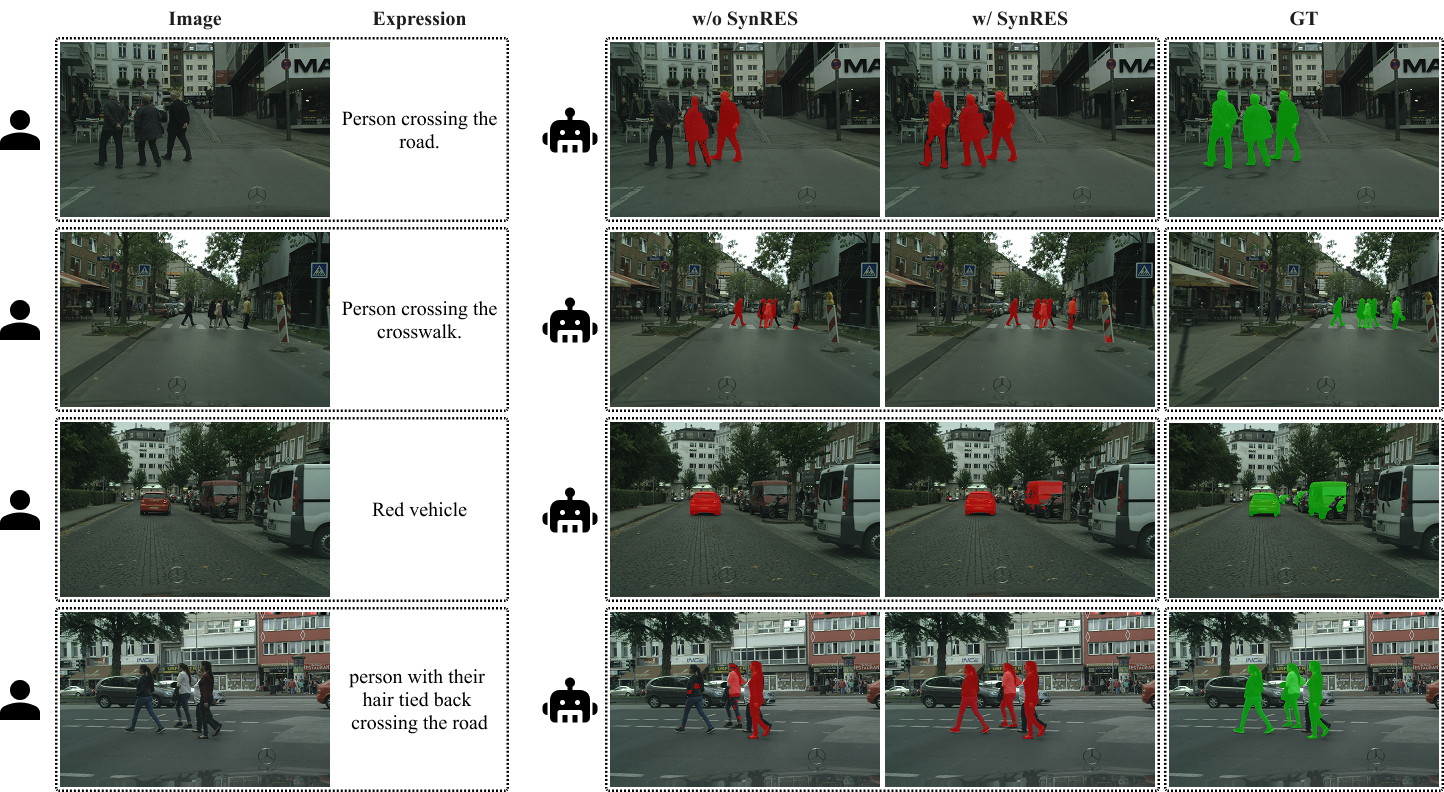}
        \caption{Visualization of Cityscapes in WildRES-DS }
        \label{fig:supplementary_cityscapes}
    \end{subfigure}
\end{figure*}

\begin{figure*}[t]\ContinuedFloat
    \begin{subfigure}{\textwidth}
        \includegraphics[width=\textwidth]{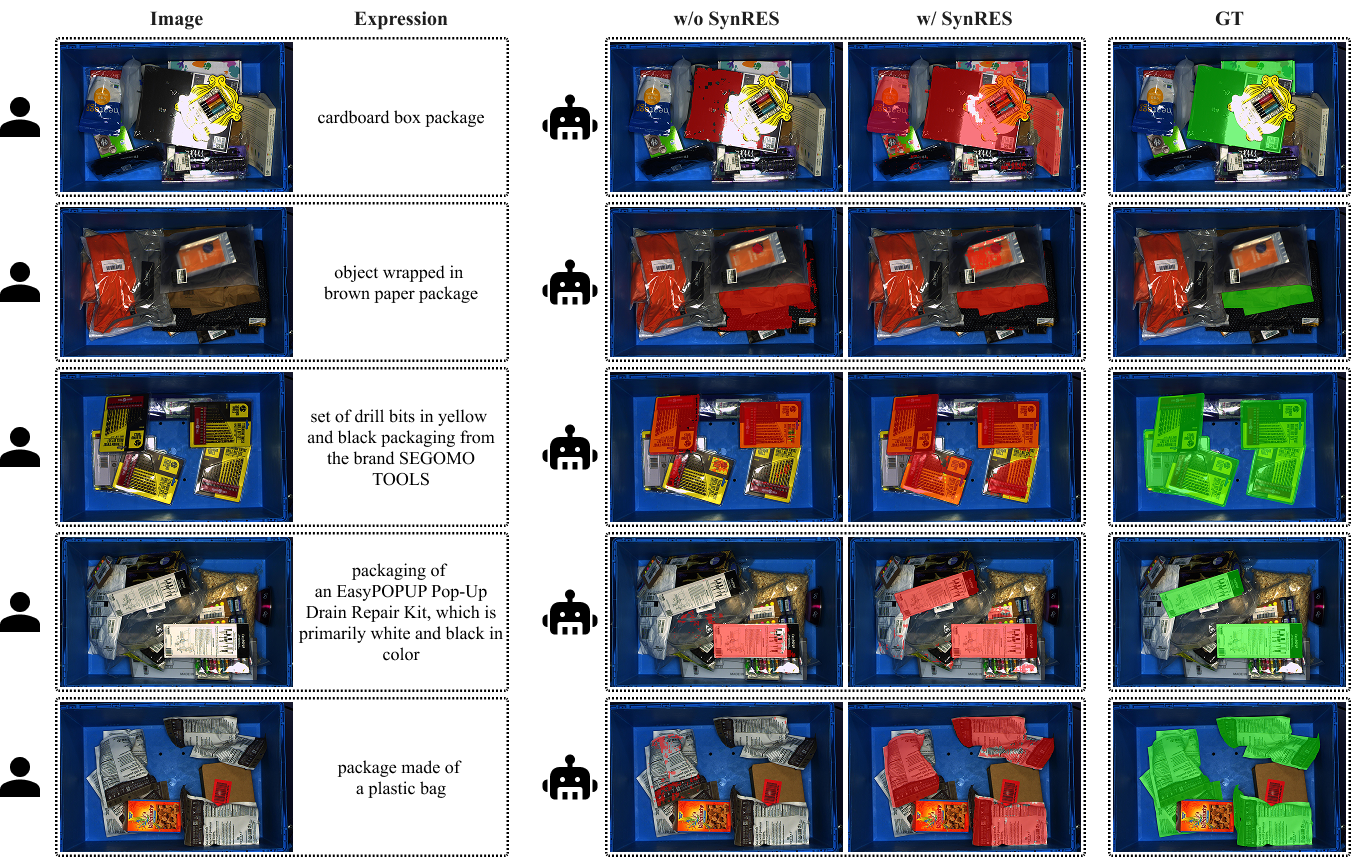}
        \caption{Visualization of ARMBench in WildRES-DS}
        \label{fig:supplementary_armbench}
    \end{subfigure}

    \caption{More sampled example from WildRES dataset.}
    \label{fig:supplementary_visualization}
\end{figure*}

\end{document}